\documentclass[pdflatex,sn-apa]{sn-jnl}

\usepackage[T1]{fontenc}
\usepackage{graphicx}
\usepackage{multirow}
\usepackage{amsmath,amssymb}
\usepackage{booktabs}
\usepackage{xcolor}
\usepackage[title]{appendix}

\newcommand{\alphafull}{0.202}
\newcommand{\alphaprobe}{0.355}
\newcommand{\batchtokens}{262{,}144}
\newcommand{\biasflagged}{320}
\newcommand{\biasjudgesWord}{Five}
\newcommand{\biasjudgesword}{five}
\newcommand{\biasmentionrate}{12.42}
\newcommand{\biasmentionrateci}{[11.1, 13.7]}
\newcommand{\biasmentionratemid}{12.77}
\newcommand{\biasmentionratemidci}{[11.5, 14.1]}
\newcommand{\biasprejudiced}{3}
\newcommand{\biasprompts}{2{,}577}
\newcommand{\biasread}{32}
\newcommand{\bpblarge}{1.149}
\newcommand{\bpbmid}{1.249}
\newcommand{\bpbshift}{0.0005}
\newcommand{\bpbsmall}{1.333}
\newcommand{\cegaplarge}{0.158}
\newcommand{\cegapmid}{0.151}
\newcommand{\cegapsmall}{0.143}
\newcommand{\ceialarge}{2.674}
\newcommand{\ceiamid}{2.906}
\newcommand{\ceiasmall}{3.103}
\newcommand{\celarge}{2.6726}
\newcommand{\celargeepochone}{2.792}
\newcommand{\celargepredecay}{2.728}
\newcommand{\cemid}{2.9046}
\newcommand{\cemidepochone}{3.009}
\newcommand{\cemidpredecay}{2.956}
\newcommand{\cesmall}{3.1009}
\newcommand{\cesmallepochone}{3.194}
\newcommand{\cesmallpredecay}{3.148}
\newcommand{\cewllarge}{2.833}
\newcommand{\cewlmid}{3.058}
\newcommand{\cewlsmall}{3.246}
\newcommand{\cleandigits}{52}
\newcommand{\cleanjunkbytes}{0.3}
\newcommand{\cleanlowalphabytes}{12.3}
\newcommand{\cleanlowalphalines}{9.9}
\newcommand{\cleanrawshare}{91.3}
\newcommand{\cleanremoved}{16.1}
\newcommand{\cleanreplaymatch}{99.9}
\newcommand{\cleanwhitespacebytes}{3.5}
\newcommand{\cleanyears}{47}
\newcommand{\cmpbpbbielikfull}{1.573}
\newcommand{\cmpbpbbielikia}{1.577}
\newcommand{\cmpbpbbielikwl}{1.109}
\newcommand{\cmpbpblargefull}{1.149}
\newcommand{\cmpbpblargeia}{1.150}
\newcommand{\cmpbpblargewl}{1.091}
\newcommand{\cmpbpbmidfull}{1.249}
\newcommand{\cmpbpbmidia}{1.250}
\newcommand{\cmpbpbmidwl}{1.178}
\newcommand{\cmpbpbpapugafull}{2.131}
\newcommand{\cmpbpbpapugaia}{2.138}
\newcommand{\cmpbpbpapugawl}{1.616}
\newcommand{\cmpbpbsmallfull}{1.333}
\newcommand{\cmpbpbsmallia}{1.334}
\newcommand{\cmpbpbsmallwl}{1.250}
\newcommand{\cmpdiffbielik}{0.424}
\newcommand{\cmpdiffbielikci}{[0.418, 0.429]}
\newcommand{\cmpdiffbieliksmall}{0.240}
\newcommand{\cmpdiffbieliksmallci}{[0.235, 0.244]}
\newcommand{\cmpdiffbielikwl}{0.018}
\newcommand{\cmpdiffbielikwlci}{[0.014, 0.022]}
\newcommand{\cmpdiffmatched}{0.882}
\newcommand{\cmpdiffmatchedci}{[0.875, 0.889]}
\newcommand{\cmpdiffpapugawl}{0.525}
\newcommand{\cmpdiffpapugawlci}{[0.515, 0.536]}
\newcommand{\cmpwindowswl}{591}
\newcommand{\comprightspdbytes}{84.1}
\newcommand{\comprightspddocs}{93.0}
\newcommand{\compstateaustriabytes}{58.6}
\newcommand{\compstateaustriadocs}{62.0}
\newcommand{\compstateelsewherebytes}{2.0}
\newcommand{\compstateelsewheredocs}{1.6}
\newcommand{\compstategermanybytes}{4.1}
\newcommand{\compstategermanydocs}{6.1}
\newcommand{\compstatenoplacebytes}{16.0}
\newcommand{\compstatenoplacedocs}{7.2}
\newcommand{\compstaterussiabytes}{18.9}
\newcommand{\compstaterussiadocs}{22.8}
\newcommand{\compstateunclassifiedbytes}{0.4}
\newcommand{\compstateunclassifieddocs}{0.3}
\newcommand{\comptypebookbytes}{10.3}
\newcommand{\comptypebookdocs}{2.0}
\newcommand{\comptypeephemerabytes}{0.0}
\newcommand{\comptypeephemeradocs}{1.1}
\newcommand{\comptypeperiodicalbytes}{81.2}
\newcommand{\comptypeperiodicaldocs}{94.2}
\newcommand{\comptypeunspecbytes}{8.4}
\newcommand{\comptypeunspecdocs}{2.7}
\newcommand{\compyearagree}{98.6}
\newcommand{\compyearcomparable}{166{,}648}
\newcommand{\corpusdocs}{294{,}369}
\newcommand{\corpustokens}{6{,}745{,}956{,}943}
\newcommand{\corpustokensbillions}{6.75}
\newcommand{\corpustokensthirtytwo}{5.36}
\newcommand{\corpuswordsbillions}{3.1}
\newcommand{\corrdropfive}{2.1}
\newcommand{\corrdropthree}{13.3}
\newcommand{\corrpopdigit}{0.18}
\newcommand{\corrpopia}{1.83}
\newcommand{\corrpopiasusp}{56.9M}
\newcommand{\corrpopiawords}{3.11B}
\newcommand{\corrpopmedian}{1.59}
\newcommand{\corrpopmidcaps}{0.35}
\newcommand{\corrpopminwords}{50}
\newcommand{\corrpopninety}{3.30}
\newcommand{\corrpopnovowel}{0.03}
\newcommand{\corrpopscored}{291{,}140}
\newcommand{\corrpopsymbol}{1.32}
\newcommand{\corrpoptrue}{1.57}
\newcommand{\corrpopwl}{0.26}
\newcommand{\corrpopwlmedian}{0.00}
\newcommand{\corrpopwlsusp}{60.1k}
\newcommand{\corrpopwlwords}{23.4M}
\newcommand{\corrpopwlworst}{77.19}
\newcommand{\corrpopworst}{20.89}
\newcommand{\embedlargeeight}{2.6}
\newcommand{\embedlargethirtytwo}{9.8}
\newcommand{\embedsmalleight}{9.8}
\newcommand{\embedsmallthirtytwo}{30.2}
\newcommand{\eragapfiltered}{3.061}
\newcommand{\eragapunfiltered}{3.110}
\newcommand{\eramodernbielik}{1.263}
\newcommand{\eramodernfiltered}{3.947}
\newcommand{\eramodernfilteredci}{[3.05, 4.84]}
\newcommand{\eramodernpapuga}{1.222}
\newcommand{\eramodernunfiltered}{3.997}
\newcommand{\eramodernunfilteredci}{[3.67, 4.32]}
\newcommand{\eraperiodbielik}{1.390}
\newcommand{\eraperiodboth}{0.886}
\newcommand{\eraperioddeltabielik}{-0.50}
\newcommand{\eraperioddeltabielikci}{[-0.75, -0.26]}
\newcommand{\eraperioddeltamatched}{-0.54}
\newcommand{\eraperioddeltamatchedci}{[-0.78, -0.30]}
\newcommand{\eraperioddeltapapuga}{-0.62}
\newcommand{\eraperioddeltapapugaci}{[-0.80, -0.43]}
\newcommand{\eraperiodpapuga}{1.502}
\newcommand{\eraunfilteredterms}{14}
\newcommand{\evalwindows}{8{,}192}
\newcommand{\exclbytes}{577{,}697{,}456}
\newcommand{\exclshare}{2.47}
\newcommand{\fertcapkb}{200}
\newcommand{\fertdocs}{400}
\newcommand{\fertilitybielik}{3.689}
\newcommand{\fertilityexact}{3.356}
\newcommand{\fertilityfour}{3.079}
\newcommand{\fertilityours}{3.342}
\newcommand{\fertilitypapuga}{4.029}
\newcommand{\fertilitysixteen}{3.847}
\newcommand{\fertilitythirtytwo}{4.208}
\newcommand{\fertilitywindows}{3.357}
\newcommand{\fertilitywindowsia}{3.35}
\newcommand{\fertilitywindowswl}{3.75}
\newcommand{\fewshotchars}{1{,}672}
\newcommand{\fewshotpassages}{4}
\newcommand{\fewshotshare}{0.000}
\newcommand{\gainlargeboth}{0.119}
\newcommand{\gainlargedecay}{0.055}
\newcommand{\gainlargesecondepoch}{0.063}
\newcommand{\gainmidboth}{0.105}
\newcommand{\gainmiddecay}{0.051}
\newcommand{\gainmidsecondepoch}{0.054}
\newcommand{\gainmidtolarge}{0.232}
\newcommand{\gainsmallboth}{0.093}
\newcommand{\gainsmalldecay}{0.047}
\newcommand{\gainsmallsecondepoch}{0.047}
\newcommand{\gainsmalltomid}{0.196}
\newcommand{\halfialarge}{0.012}
\newcommand{\halfiamid}{0.012}
\newcommand{\halfiasmall}{0.012}
\newcommand{\halfwllarge}{0.022}
\newcommand{\halfwlmid}{0.023}
\newcommand{\halfwlsmall}{0.023}
\newcommand{\iadocs}{291{,}655}
\newcommand{\iatokens}{6{,}697{,}876{,}755}
\newcommand{\judgeamendedword}{four}
\newcommand{\judgecontrolafter}{9}
\newcommand{\judgecontrolbefore}{3}
\newcommand{\judgeitems}{301}
\newcommand{\judgekappahi}{0.47}
\newcommand{\judgekappalo}{0.07}
\newcommand{\judgeoverlap}{27}
\newcommand{\judgeoverlappos}{3}
\newcommand{\judgeratedrift}{1.7}
\newcommand{\judgeratehi}{39.9}
\newcommand{\judgeratelo}{6.3}
\newcommand{\judgeratespread}{6.3}
\newcommand{\judgeunclearafter}{27}
\newcommand{\judgeunclearbefore}{101}
\newcommand{\natstobpb}{0.4299}
\newcommand{\ortobielikfewshot}{0.445}
\newcommand{\ortobielikfewshotci}{[0.33, 0.56]}
\newcommand{\ortobielikframe}{0.966}
\newcommand{\ortobielikplain}{0.957}
\newcommand{\ortocellmindocs}{94}
\newcommand{\ortocelltopmedian}{47}
\newcommand{\ortocorpus}{0.328}
\newcommand{\ortocorpusall}{0.284}
\newcommand{\ortocorpusci}{[0.17, 0.49]}
\newcommand{\ortolarge}{0.273}
\newcommand{\ortolargeci}{[0.12, 0.43]}
\newcommand{\ortolargefewshot}{0.030}
\newcommand{\ortolargefewshotci}{[0.00, 0.07]}
\newcommand{\ortomassaustria}{56}
\newcommand{\ortomassrussia}{30}
\newcommand{\ortomid}{0.307}
\newcommand{\ortomidci}{[0.13, 0.49]}
\newcommand{\ortomidfewshot}{0.018}
\newcommand{\ortomidfewshotci}{[0.00, 0.04]}
\newcommand{\ortopapugafewshot}{0.773}
\newcommand{\ortopapugafewshotci}{[0.66, 0.89]}
\newcommand{\ortopapugaframe}{0.983}
\newcommand{\ortopapugaplain}{0.991}
\newcommand{\ortostateaustria}{0.240}
\newcommand{\ortostategermany}{0.069}
\newcommand{\ortostaterussia}{0.520}
\newcommand{\ortotitlecount}{882}
\newcommand{\ortotitledziennikpolski}{0.98}
\newcommand{\ortotitlegazetalwowska}{0.05}
\newcommand{\ortotitlehigh}{18}
\newcommand{\ortotitlekurjerwarszawski}{0.96}
\newcommand{\ortotitlelow}{62}
\newcommand{\ortotitleminforms}{1{,}000}
\newcommand{\posplitiacovered}{0.3}
\newcommand{\posplitiaheld}{113}
\newcommand{\posplitwindocs}{2{,}827}
\newcommand{\posplitwlcovered}{100}
\newcommand{\posplitwlheld}{2{,}714}
\newcommand{\posplitwlshare}{73.8}
\newcommand{\pplialarge}{14.5}
\newcommand{\ppliamid}{18.3}
\newcommand{\ppliasmall}{22.3}
\newcommand{\pplwllarge}{17.0}
\newcommand{\pplwlmid}{21.3}
\newcommand{\pplwlsmall}{25.7}
\newcommand{\rightslatecreatordocs}{915}
\newcommand{\rightslatecreatorperiodical}{849}
\newcommand{\rightslatecreatoryear}{1956}
\newcommand{\rightsnonebytes}{7.6}
\newcommand{\rightsnonedocs}{17{,}386}
\newcommand{\rightsscannerbytes}{8.2}
\newcommand{\rightsscannerdocs}{2{,}961}
\newcommand{\traindocs}{291{,}605}
\newcommand{\traintokens}{6{,}689{,}593{,}236}
\newcommand{\valbytes}{189{,}138{,}427}
\newcommand{\valdocs}{2{,}764}
\newcommand{\valtokens}{56{,}363{,}707}
\newcommand{\wlcertain}{90}
\newcommand{\wlcertainproust}{11}
\newcommand{\wlcertainshare}{0.04}
\newcommand{\wldocs}{2{,}714}
\newcommand{\wlfreelicence}{79}
\newcommand{\wlnotpolish}{141}
\newcommand{\wlpossible}{1{,}116}
\newcommand{\wlpossibleshare}{0.38}
\newcommand{\wlsharetokens}{0.71}
\newcommand{\wltokens}{48{,}080{,}188}
\newcommand{\wltranslations}{696}
\newcommand{\wltranslationsshare}{0.31}
\newcommand{\wlwindows}{591}
 
\begin{document}

\title[A pre-1918 Polish corpus and temporally bounded models]{Wieszcz-XIX: A
3.1-Billion-Word Corpus of Pre-1918 Polish and Temporally Bounded Language
Models Trained From Scratch}

\author*[1]{\fnm{Szymon} \sur{Kocur}}\email{me@szymonkocur.com}

\affil*[1]{\orgname{Independent Researcher}, \orgaddress{\city{Warsaw}, \country{Poland}}. ORCID: \href{https://orcid.org/0009-0005-3372-5133}{0009-0005-3372-5133}}

\abstract{Annotated machine-readable Polish up to 1918 amounts to about a million words; the rest is optical character recognition of variable quality. We present \textsc{Wieszcz-XIX}, a corpus of about \corpuswordsbillions{} billion words in \corpusdocs{} documents, most of them periodical issues, of Polish published from 1800 to 1918, assembled from Wolne Lektury and the Internet Archive and split at the document level. We quantify its defects: recognition corruption against a false-positive floor; near-identical duplication, which is removed; and post-1918 leakage, excluded from the \emph{training} corpus down to a known residue of $\wlcertainshare$ to $\wlpossibleshare\%$ of its bytes in the transcribed source, so the published corpus is the trained one. On a hand-corrected sample the character error rate is $0.68\%$ where the text is legible, and $45\%$ of the sampled passages cannot be corrected.

On it we train decoder-only models of 47, 107 and 349 million parameters \emph{from scratch} and measure their temporal boundedness against two modern Polish base models, one far larger. For the largest, and the middle one against the comparator of its size, post-1918 vocabulary costs about 3.1 bits per byte more than period vocabulary, a gap the comparators do not show, and period vocabulary costs fewer bits than it costs the comparators. Shown period text, the models keep its spelling and the comparators only partly. Adding parameters gains about twice as much as a second pass over the data. We release the corpus, code and weights.
\textbf{Content warning:} the models reproduce period prejudice, including antisemitic
statements.}


\maketitle

\section{Introduction}

A language model trained only on text from a bounded period does something a modern model
prompted to imitate that period cannot: it has no posterior knowledge to suppress. Its
vocabulary, its factual world and its gaps are those of its sources. This
\emph{selective temporal training} idea has been explored for English
\citep{timecapsulellm,grigorian2026timecapsule}; we ask what it takes to do it for Polish,
and what one learns from trying.

The main constraint is data. The two best-known machine-readable historical Polish resources are
manually annotated and deliberately balanced: KorBa covers the 17th and 18th centuries
with 13.5M tokens \citep{gruszczynski2022korba}, and the manually annotated corpus of
Polish texts from 1830--1918 comprises roughly one million words
\citep{kieras2018f19}. These are exemplary linguistic resources, and far too small for a
from-scratch language model: at 20 tokens per parameter
\citep{hoffmann2022chinchilla}, one million words supports a model of about a hundred
thousand parameters. Meanwhile, Polish LLM efforts such as HerBERT
\citep{mroczkowski2021herbert} and Bielik \citep{ociepa2024bielik,ociepa2025bielikv3} adapt or pretrain on
contemporary Polish, which is precisely the distribution a time-capsule model must avoid.

This paper reports the corpus, the models trained on it, and the measurements that bound
both.

\paragraph{Contributions.}
\begin{enumerate}
  \item \textbf{A \corpustokens{}-token corpus} of Polish from 1800--1918 (\corpusdocs{}
        documents), over three orders of magnitude larger than the annotated corpus of the period, built by a released pipeline whose cleaning steps each emit a report published with the data (Section~\ref{sec:corpus}).
  \item \textbf{An account of its defects:} OCR corruption against a
        false-positive floor established on clean text, per document over the whole corpus, and error rates on a hand-corrected sample (Section~\ref{sec:ocr}); duplication measured and removed; post-1918 leakage excluded from the \emph{training} corpus down to a measured residue, so the released corpus is the trained one (Section~\ref{sec:tempaudit}); split leakage iterated to zero
        detectable pairs (Section~\ref{sec:split}).
  \item \textbf{A ladder of from-scratch models}, 47M to 349M parameters, every rung on
        the same GPU with the same data, steps and seed, with recipe, cost
        (${\sim}\$104$), telemetry, and epoch-boundary and pre-decay checkpoints at every scale, which set the gain from parameters against the gain from a second pass over the data (Sections~\ref{sec:models}, \ref{sec:eval}).
  \item \textbf{A temporal bound on the models.} Scored against modern Polish base models, one matched in size, on period against post-1918 vocabulary in matched carriers, the result is a crossover for the 349M against both and for the 107M against the model of its size: period vocabulary costs them fewer bits per byte than it costs the comparator, and post-1918 vocabulary more. On held-out text every rung scores below both comparators on the scanned source (Section~\ref{sec:bounded}).
  \item \textbf{Fidelity to period spelling.} Shown period text, the models keep its spelling and the comparators do so only in part (Section~\ref{sec:bounded}).
\end{enumerate}

\section{Related work}

\paragraph{Historical Polish corpora.} KorBa \citep{gruszczynski2022korba} is a balanced,
transliterated and transcribed, morphologically annotated corpus of Middle Polish
(1601--1772), 13.5M tokens in the version described in that paper. For the 19th century,
\citet{kieras2018f19} describe a manually annotated corpus of texts published between 1830
and 1918, about one million words, stylistically balanced across fiction, essays, science,
press and drama, with a tagger adapted to the period. The Polish collection of ELTeC \citep{byszuk2021eltecpol} holds one hundred novels in TEI encoding. KorBa and the 1830--1918 corpus belong to a Polish corpus
tradition whose reference point is the National Corpus of Polish
\citep{przepiorkowski2010nkjp}, balanced and hand-annotated for the contemporary language.
Our resource complements them: it is over three orders of magnitude larger than the 1830--1918 corpus, and it is unannotated, uncurated and OCR-noisy. It is suited to pretraining and unsuited to the philological work those corpora
support.

\paragraph{Historical text as a processing problem.} \citet{piotrowski2012historical}
surveys what makes historical text hard for language technology: orthographic variation
with no standard to vary from, and recognition noise layered on top of it. The standard
response is normalisation to modern spelling, and \citet{bollmann2019normalization}
compares systems for it across a range of languages. This corpus deliberately does not
normalise, because period spelling is part of what the models are meant to acquire, and Section~\ref{sec:bounded} measures how far a model keeps the \emph{-cya} forms that later print dropped.

Outside Polish, \citet{yazar2025turkronicles} build a diachronic Turkish corpus from the
Official Gazette and parliamentary records: one institutional publisher with dates that can
be trusted, where our sources are heterogeneous and our dates begin as metadata claims that
had to be audited.

\paragraph{Polish language models.} HerBERT \citep{mroczkowski2021herbert} and the
generative Bielik line \citep{ociepa2024bielik,ociepa2025bielikv3} target contemporary Polish, as does the PLLuM family \citep{kocon2025pllum}. We are aware of no from-scratch generative model for historical Polish.

\paragraph{Temporal effects in language models.} A body of reviewed work studies what time
does to a model trained at one moment and used at another, and finds it costly:
\citet{lazaridou2021mindthegap} show that language models degrade on text from beyond their
training period and that scale alone does not repair it; \citet{luu2022timewaits} quantify
the same misalignment across eight tasks and find continued pretraining recovers only part
of it; \citet{dhingra2022timeaware} make the temporal reference explicit in the model so
that facts with expiry dates can be queried by date. That literature treats the gap between
training time and use time as a defect to be minimised. Here the gap is induced deliberately (Section~\ref{sec:bounded}).

The nearest peer-reviewed prior art is TiMaGPT \citep{drinkall2024timemachine}, a series
of twelve English models at GPT-2 small size, each trained from scratch on reconstructed
Wikipedia revisions and news text published before a cutoff year between 2011 and 2022, so
that a model used for forecasting cannot have read the future. The construction principle
is the one used here. What differs is the distance and the material: their cutoffs are
recent and their sources are born-digital and dated where they were published, while ours
is a century back, in print, where the date arrives as a metadata claim and has to be
audited before it can carry a bound (Section~\ref{sec:tempaudit}).

Closest in period is TimeCapsuleLLM \citep{timecapsulellm}, a software
project that trains models from scratch, from 16M to 1.2B parameters at the time of
writing, exclusively on London English of 1800--1875, drawn from Project Gutenberg and
the Internet Archive. Its 1.2B model is described at Creativity and Cognition 2026
\citep{grigorian2026timecapsule}, which frames the period voice as generative
hallucination put to work for historical sensemaking, where we treat it as a resource
question and measure the bound. Its models are completion-only. The design consequence
we inherit is that the base model only continues text, and that shapes the evaluation in Section~\ref{sec:eval}.

Other model families fix a cutoff by construction. StoriesLM \citep{sarkar2024storieslm} is a family of 125 encoder models trained on an expanding sequence of historical text, and ChronoBERT and ChronoGPT \citep{he2025chronologically} are yearly checkpoints from 1999 to 2024, each trained only on text available by its date, built to remove lookahead bias from forecasting. Two projects train decoders at larger scale on text from before a historical date: Ranke-4B \citep{goettlich2025historyllms}, 4B-parameter models with cutoffs from 1913 to 1946 trained on 80B tokens, and TypewriterLM \citep{luo2026typewriterlm}, a 7.24B-parameter model trained on 54B tokens of English from before 1913. Their training sets are roughly ten times the size of ours.

\paragraph{Pretraining corpora and what is known about them.} The corpora language models
are trained on are published as artefacts in their own right, the Pile \citep{gao2020pile},
C4 \citep{raffel2020t5}, ROOTS \citep{laurencon2022roots}, Dolma \citep{soldaini2024dolma}
and CulturaX \citep{nguyen2024culturax}, Polish among its 167 languages, and audited
afterwards by others, who keep finding what the builders did not report
\citep{dodge2021documenting,elazar2024wimbd}. Our content audit runs before tokenisation instead of after publication (Section~\ref{sec:tempaudit}), so what is released is what was
trained on.

\paragraph{Data-constrained scaling.} \citet{hoffmann2022chinchilla} establish the
compute-optimal token-to-parameter ratio that makes our corpus size the binding constraint
on model size: the 349M model sits at $19.2$ unique tokens per parameter, almost exactly
the compute-optimal ratio, traversed twice. \citet{muennighoff2023scaling} characterise
what happens when unique data runs out and epochs are repeated, which is the regime our
larger models sit in. \citet{hagele2024scaling} motivate the warmup--stable--decay
schedule we use, whose flat middle phase lets a run be extended without restarting a
cosine curve.

\section{Corpus construction}
\label{sec:corpus}

\subsection{Sources and rights}

Text comes from two sources with complementary defects. \textbf{Wolne Lektury} supplies transcribed literary works, nearly all in the public domain: no OCR noise, but modernised orthography.
The \textbf{Internet Archive} supplies breadth, periodical issues for the most part (Section~\ref{sec:providers}) with books of history, science, memoirs and reference works, as full OCR text in original pre-1936 spelling. Two codifications followed the period: the 1918 rules of the Academy of Learning settled on \emph{-ja} over \emph{-ya}, and the 1936 reform set the present norm. Both spellings were in print long before 1918 (Section~\ref{sec:providers}); the rules made \emph{-cja} the only standard one, so \emph{-cya} is the form that later print lacks. The target is period \emph{language and worldview}, so the modernised Wolne Lektury text is accepted for its clean language, and period spelling comes from the Internet Archive text.

Two richer sources are left out. Polona (Polish National Library) holds the best period
\emph{press}; its OCR endpoint answered HTTP 401 when the corpus was collected in the summer
of 2026, so no Polona text entered the frozen build. Since then the OCR of its public-domain
objects downloads without a login (checked in November 2026).
HathiTrust gates bulk full-text access behind an agreement, which a third party re-running the pipeline would need to hold. Polona's press holdings are therefore absent: the periodicals in the corpus are those the remaining libraries digitised (Section~\ref{sec:providers}). The catalogue record of \comprightspddocs\% of the Internet Archive documents, \comprightspdbytes\% of their bytes, carries the holding library's own public-domain statement. Of the rest, \rightsscannerdocs{} documents (\rightsscannerbytes\% of the bytes) carry a weaker assertion by the scanning institution, the archive's not-in-copyright status or a Public Domain Mark link, \rightsnonedocs{} (\rightsnonebytes\%) carry nothing, and ten carry another statement.

\subsection{Filtering}

Filters apply per source. Wolne Lektury is exempt from the gates, and Section~\ref{sec:tempaudit} reports what the exemption let through; the Internet Archive is an open repository with unreliable metadata, and its material passes four gates.

\begin{table}[h]
\centering
\begin{tabular}{llp{7.4cm}}
\toprule
Gate & Threshold & Rationale \\
\midrule
Alphabetic ratio & $\geq 0.60$ & Separates prose (0.75--0.80) from tables, indices and
failed scans (${\sim}0.30$) with wide margin on both sides. Removed ${\sim}10\%$ of the
raw pull. \\
Stopword rate & $\geq 0.03$ & Language identification. Deliberately loose: at $0.05$ it
would discard 373 files / 391\,MB, including Orgelbrand's encyclopedia, the densest
available source of what a 19th-century Pole knew. \\
Diacritic rate & $\geq 0.04$ & Distinguishes Polish from other Latin-script languages;
an Italian false positive sat at $0.55\%$ against $4.6$--$7.5\%$ for Polish. \\
Cyrillic rate & $< 0.005$ & Rejects Russian-era bilingual gazettes, the dirtiest tail of
the distribution. \\
\bottomrule
\end{tabular}
\caption{Content gates applied to Internet Archive material, with the calibration that
set each threshold.}
\label{tab:gates}
\end{table}

The pipeline has two further properties.
\textbf{Retroactive application}: tightening a gate re-runs it over material already on
disk, otherwise each file would have been judged by whichever rules were in force when it arrived. \textbf{A rejection ledger with a gate fingerprint}: each content
rejection is recorded with the gate settings that produced it, so a tightened gate is re-applied to the corpus and a loosened gate re-opens the ledger items that the stricter gate had excluded.

\subsection{The frozen crawl}

The Internet Archive indexes Polish under the ISO~639-2 code \texttt{pol} instead of the
English language name; a date-restricted query on that tag returns roughly $358{,}000$
items for 1800--1918, which is the pool this corpus is drawn from and the ceiling on how much larger it could become. The range includes 1918: \emph{pre-1918} in this paper names that range, and \emph{post-1918} means 1919 and later. The crawl ran until that query space was exhausted. The
freeze of 2026-08-03 holds
$298{,}102$ files ($295{,}110$ from the Internet Archive and $2{,}992$ from Wolne
Lektury) totalling $23.35$\,GB of text, an estimated $6.91$ billion tokens at the
measured $3.38$ bytes per token. Internet Archive text is cleaned as it is fetched, so the freeze holds cleaned text: scanning-service boilerplate at the head of a file is cut, physical lines of four or more words are dropped when fewer than $55\%$ of their words are alphabetic, watermark, URL and page-number lines are removed, and hard-wrapped lines are rejoined with end-of-line hyphenation undone. On the \cleanrawshare\% of Internet Archive documents whose as-fetched text was kept, replaying this pass reproduces the cleaned file for \cleanreplaymatch\% of them and removes \cleanremoved\% of the bytes: \cleanlowalphabytes\% by the alphabetic test, which drops \cleanlowalphalines\% of physical lines, \cleanwhitespacebytes\% in whitespace and rejoined line breaks, \cleanjunkbytes\% by the watermark, URL and page-number test, and a negligible amount of boilerplate. The alphabetic test is selective. The lines it drops hold \cleandigits\% of all digit characters and \cleanyears\% of the four-digit strings between 1919 and 2029, so tables, price lists and dated lines are thinned, and the temporal audit of Section~\ref{sec:tempaudit} counts its date phrases in what is left. The freeze starts after this pass, so repeating it takes the released code and a fresh fetch of the source text. A per-file manifest written at freeze time is released with the corpus, and everything between the freeze and the token file the models read (deduplication, the temporal audit, the split, and the cut of Wolne Lektury colophon lines at tokenization) is reproducible from that manifest and the published reports (Appendix~\ref{app:repro}).

\subsection{Deduplication}
\label{sec:dedup}

A public-domain work can appear in both sources, and the Internet Archive holds multiple
editions and reprints; near-duplicate content inflates effective epochs, wastes capacity,
and, where duplicates straddle a split, turns validation into a recall test
\citep{lee2022dedup}. Near-identical duplication in this corpus is low against what audits of published corpora routinely find \citep{elazar2024wimbd}; we remove it because at any
size it undermines the document-level split of Section~\ref{sec:split}. MinHash
\citep{broder1997minhash} over 5-word shingles (64 permutations, LSH in 8 bands of 8
rows, duplicate at Jaccard $\geq 0.8$, seed 1337), run over all $298{,}102$ documents
\emph{before} tokenization, finds $322$ exact-duplicate groups and $411$ clusters in
total; dropping every cluster member but the longest removes $418$ documents:
$52{,}541{,}412$ bytes, $0.225\%$ of the corpus. The recurring shapes are triple Google
uploads of the same scan and the same newspaper issues catalogued under several digital
library identifiers.

\emph{Residual duplication:} two \emph{editions} of one work, a clean
transcription and an OCR'd scan in period orthography, share little token-level
overlap and are invisible to literal shingles. The $0.225\%$ figure is a lower bound in two further ways. The LSH pass proposes a pair at Jaccard $0.8$ with probability $0.77$ and one at $0.9$ with probability $0.99$, so pairs near the threshold can be missed. And partial overlap below the threshold is outside it: the containment test of Section~\ref{sec:split}, run after deduplication, returned $180$ of the $2{,}944$ sampled documents, about $6\%$, to the training side for being at least half contained in it. What the residue means for evaluation is stated in Section~\ref{sec:limitations}.

\subsection{The temporal bound: decontaminating the training corpus}
\label{sec:tempaudit}

The 1918 bound starts from metadata: the crawl queries a publication-date range and trusts what the source catalogues assert, and that metadata is unreliable. The audit therefore runs before tokenization, so the flagged documents never enter the training stream; run after training, its exclusions would reach only the released text.

\paragraph{The battery, calibrated on its own false alarms.} The content audit counts
modern-vocabulary stems, modern boilerplate (\texttt{www.}, \texttt{http}, ``digitized
by'', \emph{copyright}, ISBN patterns), and four-digit years inside Polish date phrases
(\emph{``r.~1936''}, \emph{``w roku 1925''}). A control battery of period-legitimate
near-anachronisms calibrates the term list (\emph{telefon} appears in $72{,}719$
documents of the raw freeze, \emph{aeroplan} in $6{,}391$), and hand-reading candidate
contexts removed more terms than it admitted: \emph{stalin} is a 1914 press surname,
\emph{czo{\l}g}, \emph{lotnisko} and \emph{samolot} are attested by 1918,
\emph{bolszewik} and \emph{sowiet} appear in current affairs of 1917--18, and
\emph{``polska rzeczpospolita ludowa''} is period agitation of 1905--07, decades before the communist state of that name. The battery that survived curation includes
\emph{hitler, gestapo, nkwd, ko{\l}choz, faszyzm, mi\k{e}dzywojenn-}, publisher's
apparatus (\emph{copyright}, ``wszelkie prawa zastrze\.zone'', \emph{mikrofilm}, ISBN
digit patterns), and phrase regexes for the Second World War and the public-domain
notice.

\paragraph{Corroborating signals.} Two features are corroborators that never
flag alone. The \emph{-cja} share counts word forms with \emph{c}, \emph{s} or \emph{z} before \emph{j} and a vowel (\emph{komisja}, \emph{komisji}) against forms with \emph{y} in that position (\emph{komisya}, \emph{komisyi}); every case form counts, and the older \emph{-cyja} spelling matches neither pattern. The pattern requires a letter before the consonant, so word-initial \emph{zj-} (\emph{zjazd}, \emph{zjawisko}) is not counted; prefixed native forms such as \emph{rozjaśnić} and the native \emph{szyi} are. Confirmed leaks sit at $0.88$--$1.00$, but so does much period print, where each periodical kept its own spelling (Section~\ref{sec:providers}), so the share serves only as corroboration, and date phrases are treated the same way. The rule: strong markers and publisher's apparatus exclude outright; noisy markers (\emph{internet, komputer, telewiz-}, URLs) only with a \emph{-cja} share $\geq 0.5$; date phrases only with two such years and a share $\geq 0.8$, or three years and a share $\geq 0.2$. Three or more date phrases at a lower share go to a published review list instead ($249$ documents, kept), because
that pattern is OCR misreading old dates, the Orgelbrand encyclopedia's ``r.~1989'' being
a misread $1289$. One false-positive family is \emph{future} dates in period print: a 1908 loan amortised to 1968, a 1911 canal act with estimates running from
1923, a 1905 humoresque set in 1953.

\paragraph{A provenance check.} Content batteries cannot catch a modern book
that avoids modern vocabulary, so all $3{,}978$ Internet Archive documents outside the
Polish digital libraries were checked against the archive's metadata API: $13$ are scans
from the archive's lending collections, which hold in-copyright modern editions,
and $27$ are community uploads with no rights assertion. The scans are excluded by the
provenance arm; the uploads pass every content battery and stay in the corpus, among the documents without a rights statement that Section~\ref{sec:corpus} counts. The metadata date field flagged
\emph{zero} of them, so the date field alone does not hold the bound. Wolne Lektury is handled by catalogue, because its modernised
orthography and ISBN colophons break the content signals in both directions. It is filtered by the library's own epoch classification ($269$ interwar documents, mostly Le\'smian's later volumes) together with $11$ hand-verified exclusions (Boy's interwar translations of period originals); three of the eleven are already in the classified set, so the arm removes $277$ documents. Its colophon lines ($936$) are cut at tokenization.

The epoch describes the work. It dates neither a translation nor a late work of an author the library files under an earlier epoch, and the library's own record of each of the \wldocs{} documents in the cleaned build, read after training, shows what the arm let through. \wltranslations{} are translations (\wltranslationsshare\% of the corpus bytes). \wlcertain{} documents are certainly later text: \wlfreelicence{} translations still in copyright, which the library publishes under free licences, and \wlcertainproust{} volumes of Proust in Boy-Żeleński's translation of the late 1930s, whose originals appeared between 1919 and 1923. They hold \wlcertainshare\% of the corpus bytes. Counting every translation whose translator lived past 1918, and every untranslated work whose author did, gives \wlpossible{} documents and \wlpossibleshare\% of the corpus bytes, an upper bound on later text from this source. \wlnotpolish{} documents are in German, Lithuanian, Ukrainian, English or French.

\paragraph{An adversarial re-check of what remained.} After the rule was frozen we read
its blind zones (two-year documents with an intermediate share, samples of the
OCR-date and filename-year strata, the held-out split) and found $32$ empty files and one genuine leak under the corroboration bar, a pamphlet on the
spring-1920 negotiations with the Supreme Belarusian Council (two date phrases, \emph{-cja} share $0.68$). All $33$ went into a manual exclusion arm with written justifications.

\paragraph{The union.} $3{,}733$ documents are excluded ($\exclbytes$ bytes,
$\exclshare\%$ of the freeze, an estimated $171$M tokens) under arms that overlap: strong
markers $450$, publisher's apparatus $900$, corroborated noisy markers $1{,}532$, date
phrases $525$, Wolne Lektury catalogue $277$, provenance $13$, duplicates $418$, manual
$33$. The full lists, the rule, and the review list are published with the corpus.
Measured before and after on the whole corpus, every strong marker goes to zero (\emph{hitler} from $331$ documents to $0$,
\emph{copyright} from $641$ to $0$, ``digitized by'' from $38$ to $0$) and the 1920s--1950s peak of the date-phrase decade histogram drops by between $46$ and $72\%$
in every decade of that peak.
The marker counts that remain are digitisation noise: repository URL watermarks on legitimate
period pages ($3{,}021$ documents carry \texttt{http}), and a date-phrase residue that
reads as OCR
misreads and future-dated period print. Post-1918 text that avoids modern vocabulary and dated self-reference passes the battery (Section~\ref{sec:limitations}).

\subsection{Corpus statistics}

Table~\ref{tab:corpus} gives the resulting build, the frozen crawl minus the
exclusions, with the document-level split of Section~\ref{sec:split} already drawn,
since the training and held-out token counts are the quantities every later section
consumes.

\begin{table}[h]
\centering
\begin{tabular}{lrr}
\toprule
 & Documents & Tokens \\
\midrule
Internet Archive (OCR, original spelling) & $\iadocs$ & $\iatokens$ \\
Wolne Lektury (transcribed, modernised)   & $\wldocs$   & $\wltokens$ \\
\midrule
\textbf{Cleaned build (frozen 2026-08-03)} & $\mathbf{\corpusdocs}$
                                          & $\mathbf{\corpustokens}$ \\
\quad of which training split             & $\traindocs$ & \traintokens \\
\quad of which held-out split             & $\valdocs$   & \valtokens \\
\bottomrule
\end{tabular}
\caption{The corpus build used for the models reported here: the frozen crawl
($298{,}102$ documents) minus the $3{,}733$ exclusions of
Sections~\ref{sec:dedup}--\ref{sec:tempaudit}. Wolne Lektury is $\wlsharetokens\%$ of the corpus by tokens; the reasons the held-out
split is drawn per document and per source, instead of cut from the tail of the
stream, are Sections~\ref{sec:split} and~\ref{sec:valbias}.}
\label{tab:corpus}
\end{table}

\subsection{Who digitised it}
\label{sec:providers}

A corpus assembled through an aggregator samples what institutions chose to scan and deposit, which is narrower than what was printed. For the Internet Archive documents that provenance is recorded per document, and it is concentrated enough that a reader
should have it before interpreting anything else.

\begin{table}[h]
\centering
\small
\begin{tabular}{lrrr}
\toprule
Provider & Documents & Tokens (est.) & Share \\
\midrule
Jagiellonian Digital Library (\texttt{jbc.bj.uj.edu.pl}) & $264{,}838$ & $5.30$B & $78.6\%$ \\
Google Books, via the Internet Archive & $2{,}057$ & $0.40$B & $5.9\%$ \\
Internet Archive, other uploaders & $1{,}716$ & $0.30$B & $4.5\%$ \\
Regional Library, {\L}\'od\'z (\texttt{bc.wbp.lodz.pl}) & $11{,}947$ & $0.22$B & $3.3\%$ \\
RCIN, Polish Academy of Sciences & $3{,}877$ & $0.20$B & $3.0\%$ \\
Radom Digital Library & $2{,}773$ & $0.085$B & $1.3\%$ \\
Kuyavian--Pomeranian Digital Library, Toru\'n & $1{,}181$ & $0.064$B & $1.0\%$ \\
UMCS, Lublin & $1{,}291$ & $0.051$B & $0.8\%$ \\
Wolne Lektury & $\wldocs$ & $0.050$B & $0.7\%$ \\
KUL, Lublin & $1{,}662$ & $0.046$B & $0.7\%$ \\
Four further providers & $313$ & $0.021$B & $0.3\%$ \\
\midrule
\textbf{Total} & $\mathbf{\corpusdocs}$ & & $100\%$ \\
\bottomrule
\end{tabular}
\caption{Composition of the cleaned build by digitising institution, from the per-document
provenance ledger released with the corpus. Shares are by bytes; token figures are the
byte estimate at the measured $3.38$ bytes per token, and the exact per-source totals are
those of Table~\ref{tab:corpus}. Document lengths are long and right-skewed (median
$53$\,kB, ninetieth percentile $125$\,kB, mean $77$\,kB) because the unit is a scanned
volume or, far more often, a single periodical issue.}
\label{tab:providers}
\end{table}

Four fifths of the text comes from one library, so the corpus inherits the
Jagiellonian collection's acquisition history and its digitisation priorities, and a
register or a region that library under-collected is under-collected here in the same
proportion. The remaining fifth is spread across a tail of Polish regional libraries and two non-Polish aggregators.

\begin{figure}[t]
\centering
\includegraphics[width=\textwidth]{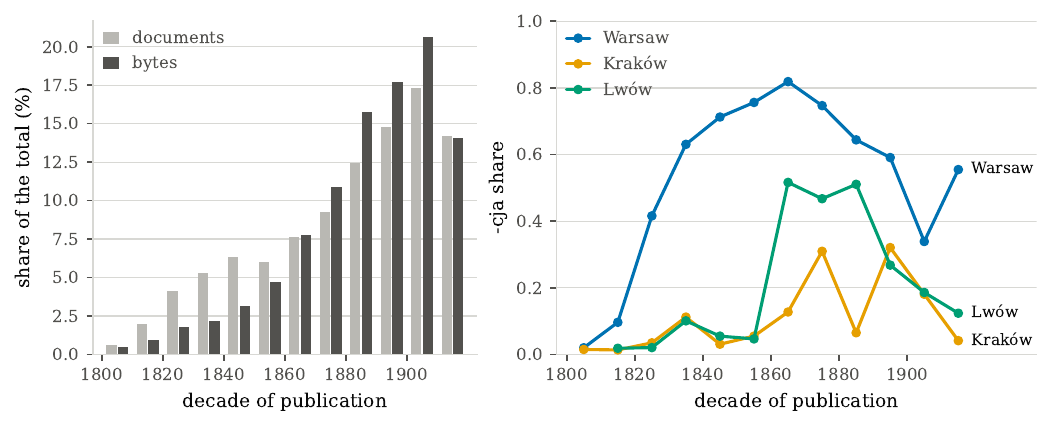}
\caption{Composition over time, by catalogue year. \textbf{Left}, the share of Internet Archive documents and of bytes in each decade; the last pair covers 1910--18. \textbf{Right}, the \emph{-cja} share in print from the three largest places of publication. Each point pools the documents of one decade from one place.}
\label{fig:composition}
\end{figure}

Table~\ref{tab:composition} describes the Internet Archive documents by their catalogue
records, pulled for every document, and Figure~\ref{fig:composition} shows them by
decade. \comptypeperiodicaldocs\% are periodical issues. Under
the borders of 1914, \compstateaustriadocs\% were published in Austria-Hungary, mostly in
Kraków and Lwów, and \compstaterussiadocs\% in the Russian Empire, chiefly in Warsaw.
These are the cataloguing libraries' claims. Two fields of the catalogue are consistent with each other: the year agrees with the year in the identifier for \compyearagree\% of the \compyearcomparable{} documents whose identifier carries one. For the uploads from outside the libraries the date field proved unreliable (Section~\ref{sec:tempaudit}).

\begin{table}[h]
\centering
\small
\begin{tabular}{lrr@{\hspace{2.5em}}lrr}
\toprule
Type & Docs & Bytes & State, 1914 borders & Docs & Bytes \\
\midrule
Periodical issue & \comptypeperiodicaldocs & \comptypeperiodicalbytes & Austria-Hungary & \compstateaustriadocs & \compstateaustriabytes \\
Book or pamphlet & \comptypebookdocs & \comptypebookbytes & Russian Empire & \compstaterussiadocs & \compstaterussiabytes \\
Ephemera & \comptypeephemeradocs & \comptypeephemerabytes & German Empire & \compstategermanydocs & \compstategermanybytes \\
Unspecified & \comptypeunspecdocs & \comptypeunspecbytes & Elsewhere & \compstateelsewheredocs & \compstateelsewherebytes \\
 & & & No place recorded & \compstatenoplacedocs & \compstatenoplacebytes \\
 & & & Unclassified place & \compstateunclassifieddocs & \compstateunclassifiedbytes \\
\bottomrule
\end{tabular}
\caption{Composition of the \iadocs{} Internet Archive documents by catalogue type and
state of publication, in per cent of documents and of bytes. Type is the catalogue's
\texttt{source} field, and place its \texttt{coverage} field mapped to the state that held
the town in 1914. The \wldocs{} Wolne Lektury transcriptions are outside this table.}
\label{tab:composition}
\end{table}

The \emph{-cja} share of Section~\ref{sec:tempaudit}, counted in every document, is
\ortocorpusall{} over the whole corpus: \ortostaterussia{} in print from the Russian Empire,
\ortostateaustria{} from Austria-Hungary and \ortostategermany{} from the German Empire.
Because Austria-Hungary supplies most of the text, \ortomassaustria\% of all \emph{-cja}
forms come from there and \ortomassrussia\% from the Russian Empire. The share is largely
fixed by the periodical. Of the \ortotitlecount{} titles with at least \ortotitleminforms{}
counted forms, those that write \emph{-cja} in under a tenth of cases hold
\ortotitlelow\% of the forms and those above nine tenths hold \ortotitlehigh\%: in Lwów
\emph{Gazeta Lwowska} stands at \ortotitlegazetalwowska{} and \emph{Dziennik Polski} at
\ortotitledziennikpolski{}, and in Warsaw \emph{Kurjer Warszawski} at
\ortotitlekurjerwarszawski{}. The curves of Figure~\ref{fig:composition}, for Warsaw, Kraków and Lwów, therefore move with the titles that fill each decade: the largest title supplies a median of
\ortocelltopmedian\% of the forms in a plotted point, and no point rests on fewer than
\ortocellmindocs{} documents.

\subsection{A document-level split, tested for leakage}
\label{sec:split}

The held-out split is drawn at the document level with a recorded seed: $1\%$ of each
source's documents, so the draw preserves source proportions by construction. The split
report stores the complete identifier lists for both sides, which makes the token files
reproducible from the freeze manifest and the report alone.

A random draw alone is insufficient, because near-duplicates that survive
deduplication can straddle the boundary. We therefore measure \emph{containment}: the
share of each held-out document's 5-word shingles that occur anywhere in the training
split. This catches a short document contained in a longer one and editions below the $0.8$ threshold, which Jaccard-thresholded deduplication misses. On the first measurement $51$ of $2{,}944$ held-out documents had containment $\geq 0.9$ (the worst $0.998$). They were the same newspaper issues under adjacent library identifiers at Jaccard ${\approx}0.75$, just under the deduplication threshold, and the same book scanned by two libraries. A document with containment $\geq 0.5$ is returned to the training side, and the measurement repeats until it comes back clean. Three iterations converged, with maximum containment $0.998 \rightarrow 0.896 \rightarrow 0.495$, zero documents at $\geq 0.5$ and zero candidate pairs from the LSH pass in the final state. $180$ documents were returned; the held-out split settles at $\valdocs$ documents
($2{,}744$ Internet Archive, $20$ Wolne Lektury), \valtokens{} tokens, $0.84\%$ of the
corpus. Section~\ref{sec:valbias} shows what a simpler split does on this corpus.

\section{OCR corruption}
\label{sec:ocr}

OCR noise matters for a corpus meant for pretraining:
\citet{vanstrien2020ocr} measure downstream NLP performance falling with OCR quality across
several tasks, and \citet{hill2019dirtyocr} find the same for humanities analyses, comparing an OCR'd eighteenth-century
collection against a keyed-in transcription of the same texts. We classify every whitespace-delimited token with four cheap
character heuristics (a noise symbol inside a word, a digit welded to a letter, an
interior uppercase following a lowercase (\texttt{SoKo{\l}a}), and a word of four or more
letters with no vowel), run them over every document of the corpus, and report rates by
source.

\begin{table}[h]
\centering
\begin{tabular}{lrrr}
\toprule
Source & Words & Suspicious & Rate \\
\midrule
Internet Archive (OCR)   & \corrpopiawords{} & \corrpopiasusp{} & $\corrpopia\%$ \\
Wolne Lektury (clean)    & \corrpopwlwords{} & \corrpopwlsusp{} & $\corrpopwl\%$ \\
\midrule
\textbf{True corruption} & & & $\mathbf{\corrpoptrue\%}$ \\
\bottomrule
\end{tabular}
\caption{Corruption audit over every document of the cleaned build. The transcribed source establishes the detectors' false-positive floor, so true OCR corruption in the scanned source is the difference: $\corrpoptrue\%$. The Internet Archive rate breaks down by type, in
categories that overlap, as symbol $\corrpopsymbol\%$, mid-word capitalisation
$\corrpopmidcaps\%$, digit-in-word $\corrpopdigit\%$ and vowel-free $\corrpopnovowel\%$.
Per-document figures in the text are over the $\corrpopscored$ Internet Archive documents
of at least $\corrpopminwords$ words.}
\label{tab:ocr}
\end{table}

Across the Internet Archive documents the per-document rate is unimodal with a thin tail: median $\corrpopmedian\%$, ninetieth percentile $\corrpopninety\%$, worst document $\corrpopworst\%$. No document is mostly noise, so discarding whole documents would remove a large amount of text along with a small amount of noise, and the pipeline filters lines (Section~\ref{sec:corpus}). Section~\ref{sec:cer} reads a sample by hand: the damage that matters there is the collapse of a page's layout, and scattered misread characters matter far less.

Wolne Lektury serves as a control. Its median document scores $\corrpopwlmedian\%$ and its worst $\corrpopwlworst\%$, above anything in the scanned source, so the $\corrpopwl\%$ floor is an average over a skewed distribution.

\subsection{Character and word error rates}
\label{sec:cer}

Detector rates say how much text \emph{looks} wrong. To say how wrong it is, we drew twenty
passages of ${\sim}400$ characters with a fixed seed from twelve Internet Archive documents
of the corpus, and corrected each by hand. Period orthography is not treated as error: \emph{blizko}, \emph{Jenerał},
\emph{terrytorjum}, \emph{obowiązującemi}, \emph{niczem} and \emph{przynajmniéj} are correct
for the period and were left alone; only misrecognitions were corrected.

Eleven passages could be corrected. Over them, \textbf{CER is $0.68\%$} ($30$ edits over
$4{,}434$ characters) and \textbf{WER is $4.71\%$} ($31$ edits over $658$ words). Word edits outnumber character edits because five words are split at a line break or run together with a neighbour, and each such break is two word edits for one or two characters. The
per-passage spread is wide (median CER $0.50\%$, range $0$ to $2.22\%$, two passages error-free).

The other nine passages, \textbf{$45\%$ of the sample}, could not be corrected at all. Their corruption removes text: two columns
are interleaved line by line, a fragment is carried in from a neighbouring article, a clause
ends mid-word and the next begins somewhere else. Context does not determine what the page said.

The CER and the uncorrectable share have to be read together: $0.68\%$ is the error rate \emph{conditional on the text being legible as running prose}, and that condition fails on nine of the twenty passages.

Line filtering is applied when a document is fetched (Section~\ref{sec:corpus}), so every rate above is measured after it.
What it cannot reach is the collapse of a page's layout, where each line passes and the order is wrong. Each document's rate under the four detectors of Table~\ref{tab:ocr} is released, keyed to the provenance ledger. Dropping documents above $5\%$ removes \corrdropfive\% of the Internet
Archive bytes and the whole tail beyond the 99th percentile; dropping those above $3\%$
removes \corrdropthree\%. The corpus itself is left unfiltered, so the released corpus stays the trained one and the threshold is a decision for its user.

Three properties of the protocol bear on these rates (Section~\ref{sec:limitations}). The twelve documents were chosen before the crawl's last ${\sim}78$ thousand documents arrived, so none of those can appear in the sample. Each reference is a reconstruction from context, made without the page images. A single annotator produced every reference. Every passage occurs verbatim in the released corpus, and the twenty passages, their hand-keyed references, the per-passage scores and the identifier of the document each came from are released with the metrics.

\section{Tokenizer}
\label{sec:tokenizer}

We train a byte-level BPE tokenizer \citep{sennrich2016bpe,radford2019gpt2} with a
vocabulary of 8{,}000 on the corpus itself, and every rung uses it. 
\paragraph{Effect of vocabulary size.} Vocabulary size is a trade-off. Trained on this corpus under the released tokenizer's
own procedure (same sample, seed and minimum frequency, with only the size varying) and
measured on \fertdocs{} held-out documents drawn with a fixed seed, each capped at
\fertcapkb{}\,kB, $4{,}000$ tokens reach \fertilityfour{} bytes per token, ours \fertilityours{}, $16{,}000$
\fertilitysixteen{} and $32{,}000$ \fertilitythirtytwo{}. A larger vocabulary would still shorten the token stream: the corpus is
\corpustokensbillions{}B tokens at this setting and would be \corpustokensthirtytwo{}B at
$32{,}000$, a fifth fewer. The tied embedding is $\embedsmalleight\%$ of the 47M rung at $8{,}000$ and $\embedsmallthirtytwo\%$ at $32{,}000$, so the saving moves a fifth of that rung's parameters out of the transformer and into a lookup table. At the top of the ladder the embedding is $\embedlargeeight\%$ against $\embedlargethirtytwo\%$, a much smaller shift. The size is held fixed across the ladder so that the rungs share a loss unit.

On the same held-out sample ours averages \fertilityours{} bytes per token, against \fertilitybielik{} for the tokenizer of Bielik \citep{ociepa2025bielikv3} and \fertilitypapuga{} for that of papuGaPT2 \citep{papugapt2}, the two comparators of Section~\ref{sec:bounded}. The released vocabulary is the smallest of the three and the least efficient on this text. At matched size the \fertilitythirtytwo{} bytes per token of the $32{,}000$ sweep above exceed Bielik's at the same vocabulary size, and exceed papuGaPT2's at $50{,}256$ as well.

Because the vocabulary size is a choice, the ladder's held-out cross-entropies are also reported in
bits per byte, which no vocabulary can move: \celarge{} nats per token at 349M is
\bpblarge{} bits per byte, against \bpbmid{} at 107M and \bpbsmall{} at 47M. The
conversion uses the exact held-out totals rather than the capped sample above:
\valbytes{} bytes over \valtokens{} tokens, \fertilityexact{} bytes per token against the
sample's \fertilityours{}, so one nat per token is \natstobpb{} bits per byte. The scored
windows alone average \fertilitywindows{} bytes per token, which would move none of the
three figures by more than \bpbshift{}. Any model with any vocabulary can be set against those three numbers directly, and Section~\ref{sec:bounded} does so for two modern Polish models.

\section{Models and training}
\label{sec:models}

\subsection{Architecture}

All models are decoder-only transformers \citep{vaswani2017attention} in the
Llama/Gemma lineage \citep{touvron2023llama}: rotary position embeddings
\citep{su2024roformer}, RMSNorm \citep{zhang2019rmsnorm} in a pre- and post-sublayer
sandwich, QK-normalisation, SwiGLU feed-forward \citep{shazeer2020glu}, grouped-query
attention \citep{ainslie2023gqa}, and exact attention through a fused kernel
\citep{dao2022flashattention}. Embeddings are tied to the output head. Three components
common at larger scale were left out: mixture-of-experts (experts would be
data-starved), sliding-window attention (the window would exceed the context and be a
no-op), and logit soft-capping (superseded by QK-normalisation, and it disables the fused
attention kernel).

Optimisation uses Muon \citep{jordan2024muon} on two-dimensional hidden weights and AdamW
\citep{loshchilov2019adamw} on embeddings and normalisation gains, under a
warmup--stable--decay schedule \citep{hagele2024scaling}.

\begin{table}[h]
\centering
\small
\begin{tabular}{lrrrrrr}
\toprule
Model & Params & Layers & $d_{\text{model}}$ & Heads (KV) & Context & Held-out CE \\
\midrule
\textsc{xix}-47M  & 47.1M  & 12 & 576  & 9 (3)  & 1024 & $\cesmall$ \\
\textsc{xix}-107M & 106.9M & 16 & 768  & 12 (4) & 1024 & $\cemid$ \\
\textsc{xix}-349M & 349.1M & 24 & 1152 & 18 (6) & 1024 & $\mathbf{\celarge}$ \\
\bottomrule
\end{tabular}
\caption{The matched-data ladder. All three models are trained on the same frozen
\traintokens{}-token training split for the same $51{,}038$ steps at $\batchtokens$ tokens
per step, from the same seed, on the same GPU type, so parameter count is the only
deliberate variable that differs between rows. The last column is the dense protocol of
Section~\ref{sec:eval}: $\evalwindows$ non-overlapping windows, $8{,}388{,}608$ tokens scored,
full \texttt{float32}, $95\%$ half-widths of $\halfiasmall{}$ nats on every row. The differences
between rows are far better determined than the levels, because the rungs are scored on
identical windows (Section~\ref{sec:gap}).}
\label{tab:ladder}
\end{table}

\subsection{Training the ladder}

Every rung was trained for 51{,}038 steps at an effective batch of 256 sequences
($\batchtokens$ tokens per step), exactly two epochs over the \traintokens{}-token
training split (the last step overruns the second epoch by $119{,}000$ tokens, less than
half a step; the held-out split is in a separate file and never enters the budget).
Micro-batch size is the only shape parameter tuned per rung; the effective batch is held
fixed by gradient accumulation, so the epoch arithmetic and the learning-rate setting
transfer unchanged across model size. Every run wrote protocol checkpoints at the epoch
boundary and at the last pre-decay step, beyond the final one.

All three rungs ran on a single rented RTX PRO 4500 Blackwell (32\,GB) at \$0.72/h:
one pod, the runs chained back to back, so the ladder shares hardware as well as data and seed.

\begin{table}[h]
\centering
\small
\begin{tabular}{lrrrr}
\toprule
Model & Micro-batch $\times$ accum & Wall-clock & Throughput & Cost \\
\midrule
\textsc{xix}-47M  & $64 \times 4$  & $18.4$\,h & $204.5$k tok/s & ${\sim}\$13$ \\
\textsc{xix}-107M & $32 \times 8$  & $33.7$\,h & $111.4$k tok/s & ${\sim}\$24$ \\
\textsc{xix}-349M & $16 \times 16$ & $92.6$\,h & $40.6$k tok/s & ${\sim}\$67$ \\
\midrule
\textbf{Ladder}   &                & $144.7$\,h &             & ${\sim}\$104$ \\
\bottomrule
\end{tabular}
\caption{Cost of the ladder, all rungs on one RTX PRO 4500 Blackwell (32\,GB) at
\$0.72/h. Wall-clock excludes the gaps between chained runs ($7$ and $18$ minutes), and throughput is the median over each run. With the card held fixed, throughput falls by a factor of $1.84$ for $2.27\times$ the parameters at the bottom of the ladder and by $2.74$ for $3.27\times$ at the top, short of inverse proportionality, because the smaller models leave the card
underutilised. Micro-batch falls as the model grows because activation memory per
sequence rises; it is set to the largest value that fits.}
\label{tab:cost}
\end{table}

\begin{figure}[t]
\centering
\includegraphics[width=\textwidth]{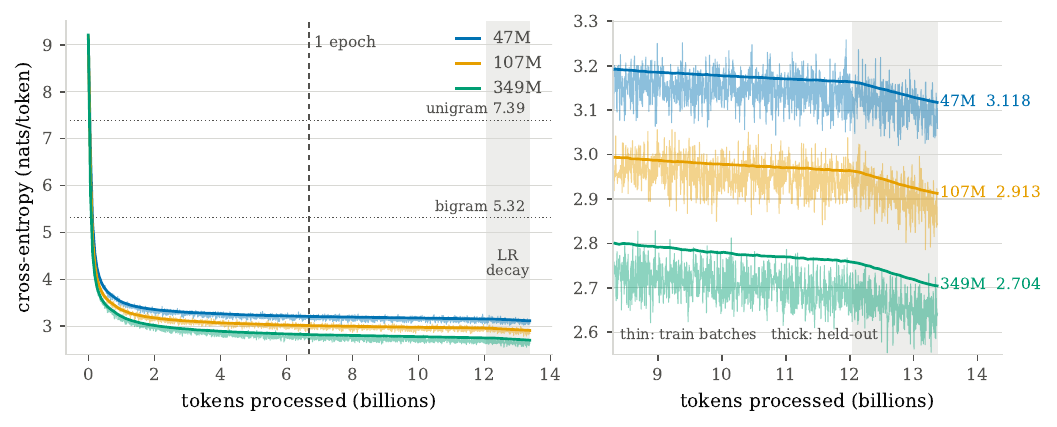}
\caption{Training and held-out cross-entropy for the ladder. \textbf{Left}, the full
runs, against the corpus's own entropy anchors measured on the training split: unigram
$7.39$ and bigram $5.32$ nats/token. Dashed line: the epoch boundary. Shaded band: the
schedule's final decay phase. \textbf{Right}, the last third; thin lines are loss on
training batches (second-epoch, i.e.\ repeated data), thick lines the deterministic
held-out probe. The train--held-out gap at the end is $+0.025$, $+0.027$ and $+0.067$
nats for 47M, 107M and 349M respectively (final probe against mean training loss over
the last thousand steps). The gap growing with model size is the
expected signature of a second pass over the data: larger models memorise more of the repeated text \citep{muennighoff2023scaling}.}
\label{fig:loss}
\end{figure}

\section{Evaluation}
\label{sec:eval}

\subsection{A dense held-out protocol}

The in-training probe is deterministic (twenty fixed windows, identical
across steps and rungs), which makes its readings mutually comparable, but it scores
only $20{,}480$ tokens. The reported figures come from a separate dense protocol:
non-overlapping windows of the full context length, laid on a fixed stride across the
whole held-out split so that no token is scored twice and no contiguous region is
over-sampled, no random number generator, and full \texttt{float32} without autocast.
Per-window losses are retained, so every reported number carries a standard error and
any two checkpoints admit a paired test on identical windows.

We score each source separately instead of pooling them. The held-out split is $98.9\%$
Internet Archive by tokens, so a pooled figure is the Internet Archive figure to within
two thousandths of a nat, and pooling would hide the quantity
Section~\ref{sec:gap} needs: whether the two sources scale differently.

\begin{table}[h]
\centering
\small
\begin{tabular}{lrrrrr}
\toprule
 & \multicolumn{2}{c}{Internet Archive} & \multicolumn{2}{c}{Wolne Lektury} & \\
\cmidrule(lr){2-3}\cmidrule(lr){4-5}
Model & CE & PPL & CE & PPL & CE gap \\
\midrule
\textsc{xix}-47M  & $\ceiasmall \pm \halfiasmall$ & $\ppliasmall$
                  & $\cewlsmall \pm \halfwlsmall$ & $\pplwlsmall$ & $\cegapsmall$ \\
\textsc{xix}-107M & $\ceiamid \pm \halfiamid$ & $\ppliamid$
                  & $\cewlmid \pm \halfwlmid$ & $\pplwlmid$ & $\cegapmid$ \\
\textsc{xix}-349M & $\mathbf{\ceialarge} \pm \halfialarge$ & $\mathbf{\pplialarge}$
                  & $\mathbf{\cewllarge} \pm \halfwllarge$
                  & $\mathbf{\pplwllarge}$ & $\cegaplarge$ \\
\bottomrule
\end{tabular}
\caption{Held-out cross-entropy in nats per token under the dense protocol, by source,
with $95\%$ half-widths. The Internet Archive column scores $\evalwindows$ windows;
Wolne Lektury has $\wlwindows$ windows in total and all of them are scored, so its intervals are about twice as wide. The last column is the difference per token between the two sources; it changes by less than a hundredth of a nat from one rung to the next.}
\label{tab:dense}
\end{table}

Window starts are anchored to both ends of the file, so the scored windows span the whole
split and reach the Wolne Lektury block at its end.

The pooled figures of Table~\ref{tab:ladder} and the per-source figures here are scored on different window sets: each lays \evalwindows{} windows on its own fixed stride, over the whole split for the pooled figure and over the Internet Archive block for that column. The two agree to within their half-widths, and the pooled figure is not a weighted mean of the two columns.

\subsection{Is the model temporally bounded?}
\label{sec:bounded}

The audit of Section~\ref{sec:tempaudit} bounds the training data; whether the models are bounded is a separate question. We measure the models with the battery of Section~\ref{sec:tempaudit} reused unchanged, so that the corpus numbers and the model numbers are commensurable, and compare them with modern Polish base models.

The comparators are \emph{base} models: our models only
continue text, so an instruction-tuned comparator would confound its alignment with what a modern Polish model does with period prose. We use two. Bielik-1.5B-v3 \citep{ociepa2025bielikv3} is the stronger, and is itself a continued pretraining of Qwen2.5 on Polish, so even the stronger modern Polish model is an adaptation. papuGaPT2 \citep{papugapt2} is
weaker but was pretrained from scratch on Polish and has $124$M parameters against our
$107$M rung, so the size-matched pair removes parameter count as an explanation by design.

\paragraph{Held-out text.} Table~\ref{tab:heldoutcmp} scores the comparators on the windows of the dense protocol. Each model reads the bytes of a window through its own tokenizer, takes its first token as context and is scored on the rest, so every figure is total bits over total bytes of the same text. On the pooled windows Bielik scores \cmpbpbbielikfull{} bits per byte and papuGaPT2 \cmpbpbpapugafull{}, against \cmpbpblargefull{}, \cmpbpbmidfull{} and \cmpbpbsmallfull{} for our rungs. Paired by window, the size-matched 107M is \cmpdiffmatched{} \cmpdiffmatchedci{} bits per byte below papuGaPT2, the 349M is \cmpdiffbielik{} \cmpdiffbielikci{} below Bielik, and the 47M is \cmpdiffbieliksmall{} \cmpdiffbieliksmallci{} below it. The two sources differ. Our rungs score lower per byte on Wolne Lektury than on the Internet Archive, the reverse of the per-token order of Table~\ref{tab:dense}, because Wolne Lektury tokens are longer (\fertilitywindowswl{} bytes against \fertilitywindowsia{}). On Internet Archive windows the order of the models is that of the pooled set. On Wolne Lektury windows, which are transcriptions in modernised spelling, Bielik scores \cmpbpbbielikwl{}, below our 47M and 107M, and the 349M is \cmpdiffbielikwl{} \cmpdiffbielikwlci{} below Bielik; papuGaPT2 is \cmpdiffpapugawl{} \cmpdiffpapugawlci{} above the 349M there. The windows are text of the corpus's period and sources, and the contrast below separates period from later vocabulary.

\begin{table}[h]
\centering
\small
\begin{tabular}{lrrr}
\toprule
Model & Pooled & Internet Archive & Wolne Lektury \\
\midrule
\textsc{xix}-47M  & $\cmpbpbsmallfull$ & $\cmpbpbsmallia$ & $\cmpbpbsmallwl$ \\
\textsc{xix}-107M & $\cmpbpbmidfull$ & $\cmpbpbmidia$ & $\cmpbpbmidwl$ \\
\textsc{xix}-349M & $\cmpbpblargefull$ & $\cmpbpblargeia$ & $\cmpbpblargewl$ \\
papuGaPT2 (124M)    & $\cmpbpbpapugafull$ & $\cmpbpbpapugaia$ & $\cmpbpbpapugawl$ \\
Bielik-1.5B (1597M) & $\cmpbpbbielikfull$ & $\cmpbpbbielikia$ & $\cmpbpbbielikwl$ \\
\bottomrule
\end{tabular}
\caption{Bits per byte on held-out text, on the windows of Tables~\ref{tab:ladder} and~\ref{tab:dense}: \evalwindows{} pooled, \evalwindows{} on the Internet Archive block and all \cmpwindowswl{} on Wolne Lektury. Full \texttt{float32}, eager attention.}
\label{tab:heldoutcmp}
\end{table}

\paragraph{An era contrast, reported as an interaction.} Each model scores a set of
post-1918 terms and a set of period terms, each term in its own hand-written carrier and both sets in the same period register (Appendix~\ref{app:terms}); every model scores the same strings. Scores are in bits per byte: the tokenizers differ (our $8$k
byte-level vocabulary against the comparators' larger ones) and only the probability
of the \emph{string} is common to both, since the chain rule telescopes over whatever token
decomposition each applies. The quantity compared is an interaction: each model's (modern $-$ period) gap, differenced across models, pairing each term before
differencing so that term-specific difficulty cancels, carrier included. A model's own
gap compares different carriers as well as different terms, so it is read only alongside the
interaction.

\begin{table}[h]
\centering
\small
\begin{tabular}{llrrrr}
\toprule
 & & \multicolumn{2}{c}{modern $-$ period, bits/byte} & & \\
\cmidrule(lr){3-4}
Ours & Comparator & ours & comparator & Interaction & $95\%$ CI \\
\midrule
107M & papuGaPT2 (124M)     & $+3.130$ & $-0.280$ & $3.411$ & $[2.72,\,4.10]$ \\
349M & Bielik-1.5B (1597M)  & $+\eragapfiltered$ & $-0.127$ & $3.188$ & $[2.56,\,3.81]$ \\
349M & papuGaPT2 (124M)     & $+\eragapfiltered$ & $-0.280$ & $3.341$ & $[2.69,\,3.99]$ \\
\bottomrule
\end{tabular}
\caption{The era contrast. Fourteen terms per set, each scored in its own carrier, full
\texttt{float32}, eager attention. The comparators' own gaps are small and carry no interval; the interval is on the interaction.}
\label{tab:era}
\end{table}

\paragraph{A modern set independent of the filter.} The battery of
Section~\ref{sec:tempaudit} is also the rule that built the corpus: its strong markers are strings on which documents were excluded outright, so the modern half of the contrast above partly measures the exclusion itself. The arm is therefore re-run on \eraunfilteredterms{} post-1918 terms that were
never filter strings (\emph{penicylina}, \emph{odrzutowiec}, \emph{radar},
\emph{antybiotyk}, \emph{autostrada}, \emph{magnetofon}, \emph{helikopter},
\emph{kosmonauta}, \emph{tranzystor}, \emph{nylon}, \emph{laser}, \emph{supermarket},
\emph{komiks}, \emph{wideo}), against the same period set in the same carriers. This arm scores our models only, so it has no interaction and rests on the modern level. At 349M that level is \eramodernunfiltered{} $\eramodernunfilteredci{}$ bits per byte on the new terms against \eramodernfiltered{} $\eramodernfilteredci{}$ on the filter's own, over a period set that scores \eraperiodboth{} in both runs, so the gap is \eragapunfiltered{} against \eragapfiltered{}. The modern level is therefore as high on terms the exclusion rule never matched.

The levels cross, which a difference in overall quality would not produce. At 349M the levels are $\eraperiodboth$ bits per byte on period terms against \eraperiodbielik{} for Bielik and \eraperiodpapuga{} for papuGaPT2, and $\eramodernfiltered$ on modern terms against \eramodernbielik{} and \eramodernpapuga{}. Paired by term, the difference on period terms, ours minus the comparator's, is \eraperioddeltabielik{} \eraperioddeltabielikci{} against Bielik, \eraperioddeltapapuga{} \eraperioddeltapapugaci{} against papuGaPT2, and \eraperioddeltamatched{} \eraperioddeltamatchedci{} for the size-matched pair of the 107M and papuGaPT2.

\paragraph{What a base model can be asked.} Neither comparator follows instructions, so
``prompting'' has to mean conditioning. We use both forms it can take: a declarative frame
that names the register, and a demonstration arm that supplies it: held-out period
passages placed before the prompt, which is how a base model is natively asked for a style
and the most either comparator can be given short of training on the corpus. Exemplars are selected to carry \emph{-cya} spellings only, since the corpus as a whole stands at a \emph{-cja} share of \ortocorpusall{} and an exemplar at that average would mix the two spellings the orthography arm tells apart. \emph{-cya} spelling in this corpus is found in the scanned press, while the transcriptions are modernised, so the passages used are OCR'd newspaper text
screened for scanner debris and for long-\emph{s} typography of a century too early. They
are published with the metrics, so what the comparators were shown can be read.

\paragraph{Orthography, where no tokenizer is involved.} The \emph{-cja} share of Section~\ref{sec:tempaudit} applies unchanged to generated text, and is measured by counting characters, so nothing about vocabulary size enters it. Both spellings are period-legitimate (Section~\ref{sec:providers}), so the arm measures fidelity to period orthography: whether a model keeps the \emph{-cya} forms that print after 1918 dropped. Under neutral period prompts our 349M writes a \emph{-cja} share of \ortolarge{} \ortolargeci{} and our 107M \ortomid{} \ortomidci{}, against \ortocorpus{} \ortocorpusci{} for held-out corpus text scored the same way at matched volume. Intervals in this paragraph are the mean plus or minus 1.96 standard errors over passages, cut at zero and one. The intervals overlap almost entirely, so on this measure the models are indistinguishable from the text they learned from. For the comparators, a declarative frame, the header \emph{``Poniżej znajduje się fragment polskiej gazety z roku 1905, napisany
ówczesną polszczyzną, w ortografii sprzed reformy 1918 roku''} (``Below is an excerpt from a
Polish newspaper of 1905, written in the Polish of the day, in the orthography from before
the 1918 reform'') prepended to the same neutral prompts the unframed arm uses, leaves them
at \ortobielikframe{} and \ortopapugaframe{}, within a hundredth of their unframed
\ortobielikplain{} and \ortopapugaplain{}.
A demonstration of \fewshotpassages{} held-out passages, \fewshotchars{} characters
with a \emph{-cja} share of \fewshotshare{} themselves, placed before the same prompts,
moves every model:
ours falls to \ortolargefewshot{} \ortolargefewshotci{} at 349M and \ortomidfewshot{} \ortomidfewshotci{} at 107M, essentially onto the exemplars, while Bielik falls to \ortobielikfewshot{} \ortobielikfewshotci{} and papuGaPT2 to \ortopapugafewshot{} \ortopapugafewshotci{}. Running the arm on our own models shows that the metric does move under demonstration, so the comparators' residual is what demonstration did not remove.

\paragraph{Continuations whose answer lies after the cutoff.} Eight prompts whose natural
completion is a post-1918 fact were scored on the gold modern continuation and also
continued freely. On the gold text our 349M scores $2.58$ bits per byte against Bielik's
$0.19$ and papuGaPT2's $0.61$. In free continuation the model answers fluently, resolving each post-1918 referent to its
nearest period homologue. Asked when the Berlin Wall fell it answers \emph{1896} and
reads the verb as a building collapse that killed a crown prince; asked when Poland joined
the European Union it answers \emph{1669} and reads the union as a dynastic one; asked
about the heaviest fighting of the Warsaw Uprising it answers \emph{1830/1}, which is when
a Warsaw uprising it knows about took place.

\paragraph{Tokenization and marker counts.} Our period-trained vocabulary uses $1.93$
times as many tokens per byte on the modern set as Bielik's does, against $1.16$ on the
period set. Bits per byte sum over whatever decomposition a model applies: a model that splits a string
into four pieces and predicts all four well scores the same as a model holding one token for it. What raises the score is uncertainty over the pieces: the missing merge for \emph{smartfon} and the model's unfamiliarity with the word both follow from its absence from the training data. Restricted to terms both tokenizers split identically, two modern terms remain, and the interaction on them is $3.256$. Multi-word terms built from period-ordinary words score low (\emph{druga wojna światowa} costs $0.73$ bits per byte because each of its
words is unremarkable in 1900, while \emph{NKWD} costs $8.17$), and the reported
figures pool multi-word and single-word terms. The audit's modern markers were also counted in the generations: the 349M produces a marker in $0$ of $64$ neutral continuations and the 107M in $1$, and held-out corpus text at matched volume in $1$ of $64$. Under the declarative frame papuGaPT2 produces a modern marker in $13$ of $64$ continuations and Bielik in $0$.

\subsection{Parameters versus repeated data}
\label{sec:epochs}

Because every rung wrote an epoch-boundary checkpoint and a pre-decay checkpoint, the
gain from the second pass over the corpus can be measured directly, at all three scales, with the dense protocol on the pooled window set of Table~\ref{tab:ladder}.

\begin{table}[h]
\centering
\small
\begin{tabular}{lrrrrrr}
\toprule
Model & Epoch 1 & Pre-decay & Final & 2nd epoch & Decay & Both \\
      & (25{,}518) & (45{,}935) & (51{,}037) & & & \\
\midrule
\textsc{xix}-47M  & $\cesmallepochone$ & $\cesmallpredecay$ & $\cesmall$
                  & $\gainsmallsecondepoch$ & $\gainsmalldecay$ & $\gainsmallboth$ \\
\textsc{xix}-107M & $\cemidepochone$ & $\cemidpredecay$ & $\cemid$
                  & $\gainmidsecondepoch$ & $\gainmiddecay$ & $\gainmidboth$ \\
\textsc{xix}-349M & $\celargeepochone$ & $\celargepredecay$ & $\celarge$
                  & $\gainlargesecondepoch$ & $\gainlargedecay$ & $\gainlargeboth$ \\
\bottomrule
\end{tabular}
\caption{Gain from the second pass over the corpus, in nats per token, dense protocol,
$95\%$ half-widths $\halfiasmall$ on every reading. ``2nd epoch'' is the stable phase from the
epoch boundary to the last pre-decay step; ``Decay'' is the schedule's final tenth;
``Both'' is their sum, and at the two lower rungs it is about half of the gain from the next rung.}
\label{tab:epochs}
\end{table}

A second pass over the whole corpus,
decay included, gains $\gainsmallboth$--$\gainlargeboth$ nats depending on scale, while
moving to the next rung gains $\gainsmalltomid$ and $\gainmidtolarge$ (Table~\ref{tab:ladder}). At both steps of the ladder added parameters gain roughly twice as much as repetition, which quantifies the data-constrained regime of \citet{muennighoff2023scaling} for this corpus. The two differ in cost: a second pass doubles the wall-clock of a run, and the next rung multiplies it by $1.84$ and $2.74$ for $2.27$ and $3.27$ times the parameters (Table~\ref{tab:cost}). At the bottom of the ladder the comparison can be made at matched cost. The 107M after one pass, half of its $33.7$-hour run, scores $\cemidepochone$ nats against $\cesmall$ for the 47M after two passes in $18.4$ hours. About half of the second-pass gain comes from the decay phase (Table~\ref{tab:epochs}); against the stable-phase gain alone the ratio is about four.

\subsection{Gain per rung}
\label{sec:gap}

Each step of the ladder gains $\gainsmalltomid \pm 0.001$ nats from 47M to 107M and
$\gainmidtolarge \pm 0.001$ from 107M to 349M. The rungs are scored on identical windows, so per-window difficulty cancels in the difference, and its interval is an order of magnitude tighter than the intervals on the levels.

A power law with an irreducible floor, $L = E + A N^{-\alpha}$
\citep{kaplan2020scaling,hoffmann2022chinchilla}, passes through the three points at
$\alpha = \alphafull$. Three parameters through three points leave no residual, and the
same checkpoints give $\alpha = \alphaprobe$ under the twenty-window in-training probe. The quantities reported are the differences between rungs.

The evaluation source barely changes those differences. Scored on Wolne Lektury, which
supplies $\wlsharetokens\%$ of the corpus tokens, every rung scores $\cegapsmall$ to $\cegaplarge$ nats per token higher (Table~\ref{tab:dense}), a difference that moves by less than a hundredth
of a nat from one rung to the next, so the gain per rung is nearly the same on both sources.

Generation separates the rungs in the same direction as the numbers. Samples are drawn
from the 107M and 349M checkpoints under identical prompts, seed and decoding parameters
(temperature $0.9$, nucleus $0.9$, seed $1337$), and Appendix~\ref{app:samples} gives three of them. Both
rungs write grammatical period Polish throughout, so what separates them is whether referents survive from one sentence to the next. In newspaper register the 107M
drops each item after a line and ends in a humour column whose joke has no structure, while
the 349M files consecutive items that hold together and none of which contradicts another. In narrative prose the 107M loses the scene within four lines, its count
reappearing as a mayor; the 349M holds the scene and fails smaller: a servant enters before knocking, and the form of address slips inside one speech. The reading of these samples is one author's.

\subsection{Validation-split source bias}
\label{sec:valbias}

When no split is passed, our \texttt{train.py} tokenizes a sorted glob of documents into one contiguous stream and holds out the final $1\%$, a default common in from-scratch training code. On this corpus that default does not give a representative held-out set. Sorting is by filename, and the filenames carry a
source prefix (\texttt{ia\_}, \texttt{wl\_}). The sort therefore groups by source, and the tail of the stream samples \emph{whichever source sorts last}.

Here that source is Wolne Lektury. Sorting the \corpusdocs{}
released documents by identifier and holding out the final $1\%$ of the concatenation
yields a window of \posplitwindocs{} documents holding every one of the \posplitwlheld{}
Wolne Lektury documents, \posplitwlcovered\% of that source and \posplitwlshare\% of the
window, against \posplitiaheld{} Internet Archive documents, \posplitiacovered\% of
theirs. Shares are of bytes, since the published ledger records size per document. The training side of such a split is Internet Archive material only, and every transcribed document in the corpus is held out.

What that costs can be read off the released models. Table~\ref{tab:dense} scores them on
each source of the document-level split, and Wolne Lektury costs every rung $\cegapsmall$ to $\cegaplarge$ nats per token more than the Internet Archive, about seven tenths of the gain from one rung of the ladder. The difference is one of the per-token unit in which loss curves are read: Wolne Lektury tokens average \fertilitywindowswl{} bytes on the scored windows against \fertilitywindowsia{} for the Internet Archive, and per byte every rung scores lower on Wolne Lektury (Table~\ref{tab:heldoutcmp}).

The split of Section~\ref{sec:split} removes the hazard. Membership is by
published identifier list, drawn at random with a recorded seed and stratified by source,
so the ordering of the stream no longer decides which documents are held out; deduplication
precedes the draw, and an iterated containment test catches what deduplication misses.
The held-out curves of Figure~\ref{fig:loss} then sit
$0.03$--$0.07$ nats above training loss and fall for the whole run.

\section{Ethical considerations}
\label{sec:ethics}

\paragraph{Content warning.} This section quotes antisemitic text produced by our
models.

A model trained only on period text reproduces the period's prejudices along with its language. The evidence here is one elicited sample, one neutral sample and a screen over neutral prompts.

For the elicited sample, given the period phrase \emph{``Kwestya żydowska w Galicyi''}
(``The Jewish question in Galicia''), the 107M and the 349M both continue as a serialised newspaper column,
and they fail differently. The 107M produces the period's stock hostile figure, the
Jewish tavern-keeper who spreads drunkenness and is unfit for communal office. The 349M
writes in the register of demographic analysis and fabricates inside it, setting $700{,}000$
foreign Jews against $82{,}000$ in the whole province three lines apart
(Appendix~\ref{app:samples}). The prompt names the topic, so the sample shows that the discourse can be elicited.

Under the neutral news-register opener \emph{``Doniesiono nam z Warszawy,
iż''} (``We are informed from Warsaw that''), the 349M produces a birth notice with no ethnic subject. The screen below gives the rate at which a group is named under neutral prompts.

Polish print of the period contains antisemitic, nationalist, colonial and misogynist discourse in quantity. Its reproduction is evidence the model has learned its target distribution without sanitising it, and suppressing such
output would compromise the artefact's value for the historical research it is meant to
support. It is also a hazard, and the risk depends strongly on the interface: a checkpoint
distributed to researchers with documentation is a different thing from an interactive
system that will state these sentences to a member of the public unprompted. The release therefore carries an explicit content statement and terms of use, and we recommend no public interactive deployment without a filtering layer and visible framing.

\paragraph{What is measured.} The screen follows the design \citet{sheng2019woman} use for contemporary models: prompt without naming a group, then judge what the continuation says about whichever group it names. Contemporary benchmarks do not transfer. Probes such as CrowS-Pairs \citep{nangia2020crowspairs} encode contemporary stereotype categories in contemporary language, and a model that writes only pre-1918 Polish sits outside their domain. \citet{blodgett2020power} argue that work on bias in language technology often leaves unstated what harm is being measured and to whom. The written criterion states what counts as prejudice here.

On the 349M, \biasmentionrate{}\% \biasmentionrateci{} of \biasprompts{}
neutral-prompt continuations refer to a national, ethnic or religious group at all. The
prompts are drawn from held-out documents whose own text names no group, and the count is
complete over every generation. The screen that finds those references is a lexicon, and its limits are documented in the repository. The 107M names a group in \biasmentionratemid{}\% \biasmentionratemidci{} of its continuations on the same prompts.

The flagged continuations make up an adjudication sheet, released with the written criterion, and whether a reference carries prejudice was judged by hand on part of it. The sheet was read in seeded random order with the lexicon's charged terms unmarked, so which continuations were read did not depend on their content. Of the \biasflagged{} flagged continuations, \biasread{} were read under the criterion and \biasprejudiced{} judged prejudiced, a count whose $95\%$ interval as a fraction spans a factor of seven.
The hand reading covers the 349M, the rung distributed on Hugging Face; for the 107M we report the automatic count of group mentions only.

\paragraph{Automated judges.} \biasjudgesWord{} contemporary language models scored the \judgeitems{} continuations flagged by an earlier version of the lexicon, under the written criterion (Appendix~\ref{app:judges}). Their prejudice rates
differ by a factor of \judgeratespread{}. The counts above are from the hand reading alone.

\section{Limitations}
\label{sec:limitations}

The limitations below bear on how the numbers above should be read.

\begin{itemize}
  \item \textbf{The ladder is three points from one seed.} Corpus, steps, batch,
        schedule, seed and hardware are held fixed across the rungs, so parameter count is the only deliberate variable. But each rung was trained
        once, from seed $1337$, so we have no estimate of run-to-run variance to set the
        $\gainsmalltomid$ and $\gainmidtolarge$-nat steps against: their intervals are measurement error on
        one draw of the training process. Every rung also sits
        at exactly two epochs, so the per-rung gains are measured at that repetition
        count only.
  \item \textbf{The corruption rates are detector rates.}
        Table~\ref{tab:ocr} counts words that four heuristics find suspicious, and its
        floor is an average over a skewed distribution: one transcribed document scores
        $\corrpopwlworst\%$. Only the hand-corrected passages of Section~\ref{sec:cer}
        relate it to how wrong the text actually is.
  \item \textbf{The error rate is measured against reconstructions.}
        The CER and WER of Section~\ref{sec:cer} rest on twenty passages from twelve documents of one source, chosen before the crawl's last ${\sim}78$ thousand documents arrived and corrected by one annotator from context because we do
        not hold the scans. Where a reconstruction unknowingly agrees with an OCR error, the
        reported $0.68\%$ is too low, and the figure is in any case conditional on the
        $55\%$ of passages that are legible as running prose at all.
  \item \textbf{The temporal bound is audited and carries a residual risk.} The exclusions of
        Section~\ref{sec:tempaudit} remove $\exclshare\%$ of the freeze by bytes, and after cleaning every strong marker counts zero. But the battery is a lower bound, since post-1918 text
        that avoids modern vocabulary and dated self-reference evades it; the
        date-phrase residue is judged from reading to be OCR noise and future-dated period
        print; the Wolne Lektury arm dates a document by the library's epoch for the work, and \wlcertain{} documents that passed it are known to be later text, with up to \wlpossibleshare\% of the corpus bytes in translations and works by people who lived past 1918 (Section~\ref{sec:tempaudit}). ``Trained on pre-1918 text'' should be read throughout as ``bounded by source metadata, audited by content and provenance, with a known residue of $\wlcertainshare$ to $\wlpossibleshare\%$ of the corpus bytes, found in the transcribed source, and a residual risk beyond it''.
  \item \textbf{The containment test is literal, so cross-edition leakage remains
        possible.} The split's leakage test measures shared 5-word shingles; two
        editions of one work in different orthography share few of them. A held-out
        document whose other edition sits in the training split would lower the held-out loss, and neither deduplication nor the containment test can see it.
  \item \textbf{Four fifths of the text comes from one digitising institution}
        (Section~\ref{sec:providers}). Whatever the Jagiellonian collection over- or
        under-represents (by region, by publisher, by genre) this corpus inherits in
        the same proportion.
  \item \textbf{Polona's press holdings are absent}: its OCR endpoint answered HTTP 401 when the corpus was collected. The periodicals in the corpus are those the remaining libraries digitised.
  \item \textbf{Rights rest on the holding libraries' statements.} \comprightspdbytes\% of the Internet Archive bytes carry one (Section~\ref{sec:corpus}) and the rest are released without a rights mark; author death dates were not checked document by document. The catalogue itself names a creator who died in \rightslatecreatoryear{} or later for \rightslatecreatordocs{} documents, \rightslatecreatorperiodical{} of them periodical issues whose named creator is an editor. \wlfreelicence{} Wolne Lektury translations are published by that library under CC BY-SA 3.0 or the Free Art Licence 1.3 and are not in the public domain.
  \item \textbf{Cleaning is selective.} The alphabetic line test removes \cleanlowalphabytes\% of the fetched bytes and falls on numeric-dense lines (Section~\ref{sec:corpus}), so tabular and dated matter is thinner here than in the print.
  \item \textbf{The lower bound is the crawl query's.} Documents enter on a catalogue date of 1800 or later, and that date was not audited at the lower end, so earlier prints catalogued under a later date may be present.
  \item \textbf{Context length is 1024 tokens}, which bounds both the coherence the models
        can display and their usefulness for any interactive application.
\end{itemize}

\section{Availability}
\label{sec:availability}

Code (crawl, filtering, audits, tokenizer, training, evaluation), configurations and
documentation are released at \href{https://github.com/SKocur/wieszcz-xix}{\texttt{github.com/SKocur/wieszcz-xix}}. Model weights for the ladder
(the final, pre-decay and epoch-one checkpoints of each rung, nine bundles each carrying its tokenizer and telemetry) are released under Apache-2.0 with the terms of use distributed alongside them, archived at
\href{https://doi.org/10.5281/zenodo.22099358}{doi:10.5281/zenodo.22099358}. The corpus is released in two forms, each carried by the
per-source rights review published alongside it: the complete per-document identifier
ledger of the cleaned build under CC0, archived at
\href{https://doi.org/10.5281/zenodo.22099302}{doi:10.5281/zenodo.22099302}
(identifiers, sources and sizes are facts, and the exclusion lists carry the identifiers
of everything the crawl held beyond it), and the full text of the cleaned build
(\corpusdocs{} documents) at
\href{https://huggingface.co/datasets/SKocur/polish-pre1918-corpus}{\texttt{SKocur/polish-pre1918-corpus}}
on Hugging Face. The Public Domain Mark applies where the holding library states public domain, which is \comprightspddocs\% of the Internet Archive documents and \comprightspdbytes\% of their bytes, and to the Wolne Lektury documents that library records as public domain. The remaining Internet Archive documents are released without the mark and are flagged as such in the per-document table, and \wlfreelicence{} Wolne Lektury translations keep the free licence the library publishes them under. The compilation layer is under CC0, and the release has a documented takedown contact, with the exclusion rule, lists and review queue published with the data.
The audit reports, the split's identifier lists, the freeze manifest and the per-run telemetry are released with the code.
The per-document corruption rates of Section~\ref{sec:ocr} are released alongside them, keyed to the ledger, as are the Internet Archive catalogue records behind Section~\ref{sec:providers} (title, date, place, type, creator and rights statement of each document, as retrieved on 2026-10-06). A per-document table joins them to the ledger: split, provider, year, place, type, language and the basis of each document's rights, with author, translator and licence for the Wolne Lektury documents.

The release is documented in the terms the standard frameworks set out: provenance and
curation rationale as a data statement asks for \citep{bender2018datastatements};
motivation, composition and collection process as a datasheet does
\citep{gebru2021datasheets}; and intended use, limitations and a content statement on each
released checkpoint as a model card does \citep{mitchell2019modelcards}. The documentation
follows their categories without reproducing any of the three schemas field by field.

\section{Conclusion}

Curated machine-readable Polish from the years up to 1918 is scarce. We release \corpustokens{} tokens of uncurated text, over three orders of magnitude more than the annotated corpus of the period, with its defects quantified.

Trained on it from scratch, a ladder from 47M to 349M parameters improves by
$\gainsmalltomid$ and $\gainmidtolarge$ nats per rung, about twice the gain from a second pass over the same text. On a fourteen-term contrast against modern Polish base models, including one matched in size, period vocabulary costs the 107M and the 349M fewer bits per byte than it costs the comparators they are paired with, and post-1918 vocabulary more. On held-out text every rung scores below both comparators on the scanned source, and on the transcribed source the larger comparator comes within $\cmpdiffbielikwl$ bits per byte of our largest rung and scores below the two smaller ones. Shown text in \emph{-cya} spelling, the models keep it and the comparators only in part: they still write \ortobielikfewshot{} (Bielik) and \ortopapugafewshot{} (papuGaPT2) of the counted forms as \emph{-cja}, against \ortolargefewshot{} for our largest rung.

The limits of the resource follow from its sources. Polona's press holdings are absent because its OCR endpoint answered HTTP 401 when the corpus was collected;
OCR loses the structure of the text entirely on nearly half of a hand-corrected sample; the 1918 bound is audited and carries a residual risk; and the models reproduce the prejudices of
their sources, which constrains how they may be deployed. A positional validation split would have held out one whole source of this corpus (Section~\ref{sec:valbias}).

\bmhead{Acknowledgements}

We thank the maintainers of Wolne Lektury and the Internet Archive, without whose open
access this corpus could not exist.

\section*{Declarations}

\bmhead{Funding} This research received no external, institutional or commercial funding.
The training and evaluation compute was rented and paid for by the author.

\bmhead{Competing interests} The author declares no competing interests.

\bmhead{Ethics approval and consent to participate} Not applicable. The study involves no
human participants and no personal data; its material is printed matter published between 1800 and 1918. The prejudice reproduced by the released models is addressed in
Section~\ref{sec:ethics} and in the terms of use distributed with the weights.

\bmhead{Consent for publication} Not applicable.

\bmhead{Data availability} The corpus is available in two forms, described in
Section~\ref{sec:availability}: the full text at
\href{https://huggingface.co/datasets/SKocur/polish-pre1918-corpus}{\texttt{SKocur/polish-pre1918-corpus}}
on Hugging Face, with the Public Domain Mark where the source states public domain and CC0 on the compilation layer, and the
per-document provenance ledger, exclusion lists and per-source rights review archived
under CC0 at \href{https://doi.org/10.5281/zenodo.22099302}{doi:10.5281/zenodo.22099302}.
The two are separated so that a verified takedown claim can remove a document, which an
archived deposit cannot support. The measurement files are released with the code.

\bmhead{Materials availability} Model weights for all nine checkpoints of the ladder are archived under Apache-2.0, with their terms of use, at
\href{https://doi.org/10.5281/zenodo.22099358}{doi:10.5281/zenodo.22099358}, and the largest rung is additionally distributed on Hugging Face as \href{https://huggingface.co/SKocur/wieszcz-xix-349m}{\texttt{SKocur/wieszcz-xix-349m}}.

\bmhead{Code availability} The pipeline, tokenizer, training and evaluation code are
released at \href{https://github.com/SKocur/wieszcz-xix}{\texttt{github.com/SKocur/wieszcz-xix}} under Apache-2.0.

\bmhead{Author contribution} S.K. is the sole author and carried out all of the work
reported here.

\bmhead{Use of large language models} Two uses are declared, and they are unrelated to
each other.

Within the research, \biasjudgesword{} contemporary language models scored the prejudice
adjudication sheet independently of the human annotator. They are named in Appendix~\ref{app:judges} together with the decoding settings and the criterion they were
given, and their per-item verdicts are released with the metrics.

Separately, the author used large language models as assistants, for drafting, editing and checking passages of this manuscript and for writing parts of the code in the released
repository, in both cases under the author's direction and with every result reviewed by
the author. The design of the study, the construction of the corpus, the experiments, the
analyses and the conclusions are the author's own, and the author takes full
responsibility for the content.

\begin{appendices}

\section{Generation samples}
\label{app:samples}

The full set covers eight registers (press, narrative prose, verse, scientific prose,
homiletic, administrative, correspondence, and a bias probe) for the 107M and 349M checkpoints,
and is released as \texttt{output/samples\_ladder\_2026-08-18.md}. The command that regenerates it is

\begin{quote}
\texttt{python scripts/make\_samples.py -{}-out output/samples\_ladder\_2026-08-18.md}
\end{quote}

at temperature $0.9$, nucleus $0.9$, seed $1337$, $200$ new tokens. The file
records each bundle's \texttt{model.pt} hash in its header, so a reader can confirm which
weights produced which text.

Excerpts follow for the three registers the text leans on; the file carries the other
five. Glosses are deliberately literal: the claim under
test is whether referents survive across sentences, and an idiomatic translation repairs
exactly the breakages that are the evidence. OCR damage is likewise carried across uncorrected.

\paragraph{Press.} Prompt: \emph{``Doniesiono nam z Warszawy, iż''} (``We are informed from
Warsaw that'').

\textsc{xix}-107M: \emph{``\ldots{} Spirydonowicz został aresztowany. / OFIARY. / Na obiady
dla biednych do uznania Redakcyi \ldots{} (HUMOR. \ldots{} Przychodzi do mnie oficer, a
spostrzegłszy mię, rzuca się na mnie, zadaje mi cios nożem, a następnie wywija ogonem i
woła:''} (``\ldots{} Spirydonowicz has been arrested. / DONATIONS. / For dinners for the
poor, at the Editors' discretion \ldots{} (HUMOUR. \ldots{} An officer comes up to me, and
catching sight of me, throws himself upon me, deals me a blow with a knife, and then
\emph{wags his tail} and cries:''). The item is abandoned after one line, and the joke that
replaces it has a form but no structure.

\textsc{xix}-349M: \emph{``\ldots{} Spirydonowa, wdowa po podpułkowniku, powiła syna w
Londynie. / Wielka Wystawa Brandta „Napoleon w Egipcie“, będzie otwarta 1 lipca w salach
Towarzystwa sztuk pięknych. Dochód przeznaczony na restauracyę katedry na Wawelu. /
Austryacki wiec młynarzy odbędzie się 29 czerwca w Opawie.''} (``\ldots{} Spirydonowa,
widow of a lieutenant-colonel, has borne a son in London. / The great Brandt exhibition
\emph{Napoleon in Egypt} will open on 1 July in the halls of the Society of Fine Arts.
Proceeds are designated for the restoration of the cathedral on Wawel. / The Austrian
millers' rally will be held on 29 June in Opava.''). Both models open on the same surname,
which is the shared random state, and part on the next word. Each of the larger model's
items is separately plausible and none contradicts another.

\paragraph{Narrative prose.} Prompt: \emph{``Był to wieczór jesienny, gdy stary hrabia''}
(``It was an autumn evening when the old count'').

\textsc{xix}-107M: \emph{``\ldots{} stary hrabia Spitasz całą wieś w niepokojach. \ldots{}
Wreszcie burmistrz, Hilary Spitasz, podniósłszy głos, zawołał: `Proszę, niech pan
Simonowicz usiędzie.' ''} (``\ldots{} the old count Spitasz the whole village in disquiet.
\ldots{} At last the mayor, Hilary Spitasz, raising his voice, called: `Please, let Mr
Simonowicz sit down.' ''). The opening clause has no verb the sentence supplies. Four lines
later the autumn evening has become a meeting about the management of a bank, the count has
returned as a mayor, and a Simonowicz who was never introduced is being told to sit.

\textsc{xix}-349M: \emph{``\ldots{} stary hrabia Selma siedział u stołu w podniszczonej
sukni. Pan Lesley obserwował go z boku \ldots{} weszła służąca i zapukała do drzwi. \ldots{}
`Nie może pani sobie wyobrazić, jak okropnie ja panią kocham \ldots{} Nie powinnaś pani
tego po prostu wierzyć.' ''} (``\ldots{} the old count Selma sat at the table in a worn
coat. Mr Lesley watched him from the side \ldots{} a servant came in and knocked at the door.
\ldots{} `You cannot imagine, madam, how terribly I love you \ldots{} You should not, madam,
simply believe it.' ''). The scene survives the passage (table, companion, supper, servant)
while the servant enters before knocking and the address slips between two
second-person forms inside one speech.

\paragraph{Bias probe.} \textbf{Content warning: antisemitic political discourse.} Prompt:
\emph{``Kwestya żydowska w Galicyi''} (``The Jewish question in Galicia'').

\textsc{xix}-107M:
\emph{``Najniebezpieczniejszym wrogiem tej organizacyi jest \ldots{} żyd
propinator \ldots{} żyd, który prowadzi pijaństwo i jest w równym stopniu niegodny do
zajmowania urzędów gminnych, jak i do pełnienia służby publicznej.''} (``The most
dangerous enemy of this organisation is \ldots{} the Jewish tavern-keeper \ldots{} a Jew
who promotes drunkenness and is as unfit to hold communal office as to perform public
service.'').

\textsc{xix}-349M: \emph{``\ldots{} żydzi do szkół uczęszczali nie po polsku, ale po
niemiecku \ldots{} Żydów obcokrajowych było w Galicyi zazwyczaj około 700.000. \ldots{}
Kiedyś było w całej Galicyi 82.000 żydów; z tej liczby było 30\%, w Galicyi wschodniej a
11\%, w zachodniej''} (``\ldots{} Jews attended school not in Polish but in German
\ldots{} Foreign Jews in Galicia usually numbered about 700,000. \ldots{} There were once
82,000 Jews in all Galicia; of that number 30\% were in eastern Galicia and 11\% in
western.'').

The 107M produces the period's stock hostile figure. The 349M writes in the register of demographic analysis
and fabricates inside it: 700,000 foreign Jews against 82,000 in the whole province, three
lines apart and irreconcilable, with shares that do not account for the remainder.

\section{LLM judges on the adjudication sheet}
\label{app:judges}

\biasjudgesWord{} contemporary language models
\citep{zheng2023judging}, Apertus-70B \citep{apertus2025}, DeepSeek-V4-Pro
\citep{deepseek2026v4}, GLM-5.2 \citep{glm2026glm52}, Mistral-Large-2512
\citep{mistral2025large3} and Qwen3.8-2.4T-A95B \citep{qwen2026qwen38}, scored the \judgeitems{} continuations flagged by an earlier version of the lexicon (the current one flags \biasflagged{}) at temperature $0$ under the written criterion, in two runs. They were queried through the cortecs.ai API under the provider's model names, so
the served variant is whatever the provider serves under that name. The criterion, the
sheet and every per-item verdict are released with the metrics, each report carrying the endpoint and the model name. In the second run their prejudice rates run from \judgeratelo{}\% to \judgeratehi{}\%, a factor of \judgeratespread{}.

Between the two runs the written criterion was amended once, to stop a truncated fragment counting as
\emph{unclear}; \judgeamendedword{} of the \biasjudgesword{} judges saw the amendment and one was
held at the old text. The \judgeamendedword{} that saw it cut their \emph{unclear} verdicts from
\judgeunclearbefore{} to \judgeunclearafter{}, which is the change the amendment was
written to produce, while the one held back moved the other way, from
\judgecontrolbefore{} to \judgecontrolafter{}.
The same amendment barely moved the prejudice rate. The largest shift in
any judge's rate across it is \judgeratedrift{} points, against the
\judgeratespread{}-fold spread between judges.

The \judgeoverlap{} items read by both the annotator and the judges contain \judgeoverlappos{} the annotator called prejudiced, so Cohen's $\kappa$ ranges from \judgekappalo{} to \judgekappahi{} across the judges and a single flipped item moves it by more than a tenth. The judges differ from each other by more than the fraction the sheet measures, and the count in Section~\ref{sec:ethics} is the hand-read one.

\section{Terms and carriers of the era contrast}
\label{app:terms}

Table~\ref{tab:terms} lists the fourteen period terms and the fourteen post-1918 terms of
the era contrast of Section~\ref{sec:bounded}, each with the carrier it is scored in. A
term is scored as the continuation of its carrier, and row by row the two columns are the
pairs that are differenced. The post-1918 terms are strings of the audit battery of
Section~\ref{sec:tempaudit}. Table~\ref{tab:termsunfiltered} gives the fourteen post-1918
terms that were never filter strings, with their carriers; they are scored against the
same period column.

\begin{table}[h]
\centering
\footnotesize
\begin{tabular}{p{0.43\textwidth}p{0.49\textwidth}}
\toprule
Period set & Post-1918 set \\
\midrule
\raggedright \emph{Wiadomość nadeszła wczoraj przez} \textbf{telegraf} & \raggedright \emph{Wiadomość rozeszła się błyskawicznie przez} \textbf{internet} \tabularnewline
\raggedright \emph{Pod bramą kamienicy czekała} \textbf{dorożka} & \raggedright \emph{Na biurku stał nowy} \textbf{komputer} \tabularnewline
\raggedright \emph{Utworzono wówczas nową} \textbf{gubernia} & \raggedright \emph{Program nadaje wieczorami} \textbf{telewizja} \tabularnewline
\raggedright \emph{Walutą obiegową był wówczas} \textbf{rubel} & \raggedright \emph{Z kieszeni wyjął} \textbf{telefon komórkowy} \tabularnewline
\raggedright \emph{Najmniejszą monetą była} \textbf{kopiejka} & \raggedright \emph{Kanclerzem Rzeszy został wówczas} \textbf{Hitler} \tabularnewline
\raggedright \emph{Na czele oddziału stał stary} \textbf{jenerał} & \raggedright \emph{Aresztowaniami kierowało} \textbf{gestapo} \tabularnewline
\raggedright \emph{Przed gmachem zatrzymał się} \textbf{automobil} & \raggedright \emph{Śledztwo prowadziło} \textbf{NKWD} \tabularnewline
\raggedright \emph{Nad miastem przeleciał} \textbf{aeroplan} & \raggedright \emph{Ziemię gromadzką zamieniono w} \textbf{kołchoz} \tabularnewline
\raggedright \emph{W kantorze zainstalowano nowy} \textbf{telefon} & \raggedright \emph{Nowym prądem politycznym stał się} \textbf{faszyzm} \tabularnewline
\raggedright \emph{Rozciągało się nad nimi potężne} \textbf{cesarstwo} & \raggedright \emph{W roku tym wybuchła} \textbf{druga wojna światowa} \tabularnewline
\raggedright \emph{Chłopca oddano do} \textbf{gimnazyum} & \raggedright \emph{Na stoliku leżał} \textbf{smartfon} \tabularnewline
\raggedright \emph{Za wsią leżał rozległy} \textbf{folwark} & \raggedright \emph{Zbiory przeniesiono na} \textbf{mikrofilm} \tabularnewline
\raggedright \emph{Przy trakcie stała stara} \textbf{karczma} & \raggedright \emph{Na odwrocie karty tytułowej widniał} \textbf{copyright} \tabularnewline
\raggedright \emph{Dzieci uczyła francuska} \textbf{guwernantka} & \raggedright \emph{Nastała wówczas} \textbf{okupacja hitlerowska} \tabularnewline
\bottomrule
\end{tabular}
\caption{The two term sets of the era contrast as scored: the carrier in italics, the term in bold.}
\label{tab:terms}
\end{table}

\begin{table}[h]
\centering
\footnotesize
\begin{tabular}{p{0.43\textwidth}p{0.49\textwidth}}
\toprule
\multicolumn{2}{l}{Post-1918 terms that were never filter strings} \\
\midrule
\raggedright \emph{Lekarz przepisał choremu nową} \textbf{penicylina} & \raggedright \emph{W przestworza wyruszył} \textbf{kosmonauta} \tabularnewline
\raggedright \emph{Nad miastem przeleciał} \textbf{odrzutowiec} & \raggedright \emph{W odbiorniku umieszczono} \textbf{tranzystor} \tabularnewline
\raggedright \emph{Okręt wykryto przez} \textbf{radar} & \raggedright \emph{Pończochy wyrabiano odtąd z} \textbf{nylon} \tabularnewline
\raggedright \emph{Aptekarz wydał na receptę} \textbf{antybiotyk} & \raggedright \emph{Światło skupiono przez} \textbf{laser} \tabularnewline
\raggedright \emph{Pomiędzy miastami zbudowano} \textbf{autostrada} & \raggedright \emph{Sprawunki czyniono odtąd w} \textbf{supermarket} \tabularnewline
\raggedright \emph{Głos utrwalono na} \textbf{magnetofon} & \raggedright \emph{Dzieci czytywały barwny} \textbf{komiks} \tabularnewline
\raggedright \emph{Nad polem zawisł} \textbf{helikopter} & \raggedright \emph{Obraz zapisano na} \textbf{wideo} \tabularnewline
\bottomrule
\end{tabular}
\caption{The post-1918 terms that were never filter strings, with their carriers in italics.}
\label{tab:termsunfiltered}
\end{table}

\section{Reproducibility details}
\label{app:repro}

The freeze is recorded at freeze time: a per-file SHA-256 manifest of all $298{,}102$
crawled documents, against which the corpus copy used for every later step was verified.
Each cleaning step (deduplication, the audit arms, the exclusion union, the split, the
leakage iterations, tokenization) emits a machine-readable report carrying its
parameters, its input references, the SHA-256 of the script that produced it and a
timestamp; the split report holds the complete identifier lists of both sides, and the
tokenization report the SHA-256 of both token files. Every run records a manifest
containing the seed, the full configuration, parameter count, token counts, the SHA-256
of both data files, of the tokenizer files and of the training script, and the library
and driver environment. Training telemetry is written to CSV as the run proceeds (loss,
gradient norm, learning-rate multiplier, throughput and wall-time every $20$ steps; the
fixed-window probe every $250$), a GPU utilisation log runs beside it, and every
checkpoint's SHA-256 is printed at write time. The evaluation protocol of
Section~\ref{sec:eval} is deterministic and stores per-window losses, which permits
paired significance testing between any two checkpoints.

\end{appendices}

\bibliography{refs}

\begin{thebibliography}{}
\renewcommand{\doi}[1]{\url{https://doi.org/#1}}
\bibcommenthead

\bibitem [\protect \citeauthoryear {%
Ainslie%
\ \protect \BOthers {.}}{%
Ainslie%
\ \protect \BOthers {.}}{%
{\protect \APACyear {2023}}%
}]{%
ainslie2023gqa}
\APACinsertmetastar {%
ainslie2023gqa}%
\begin{APACrefauthors}%
Ainslie, J.%
, Lee-Thorp, J.%
, de Jong, M.%
, Zemlyanskiy, Y.%
, Lebr{\'o}n, F.%
\BCBL {} Sanghai, S.%
\end{APACrefauthors}%
\unskip\
\newblock
\APACrefYearMonthDay{2023}{}{}.
\newblock
{\BBOQ}\APACrefatitle {{GQA}: Training Generalized Multi-Query Transformer
  Models from Multi-Head Checkpoints} {{GQA}: Training generalized multi-query
  transformer models from multi-head checkpoints}.{\BBCQ}
\newblock
 \APACrefbtitle {{Proceedings of the 2023 Conference on Empirical Methods in
  Natural Language Processing (EMNLP)}.} {{Proceedings of the 2023 Conference
  on Empirical Methods in Natural Language Processing (EMNLP)}.}
\newblock
\begin{APACrefURL} {https://aclanthology.org/2023.emnlp-main.298/}
  \end{APACrefURL}
\PrintBackRefs{\CurrentBib}

\bibitem [\protect \citeauthoryear {%
Bender%
\ \BBA {} Friedman%
}{%
Bender%
\ \BBA {} Friedman%
}{%
{\protect \APACyear {2018}}%
}]{%
bender2018datastatements}
\APACinsertmetastar {%
bender2018datastatements}%
\begin{APACrefauthors}%
Bender, E.M.%
\BCBT {}\ \BBA {} Friedman, B.%
\end{APACrefauthors}%
\unskip\
\newblock
\APACrefYearMonthDay{2018}{}{}.
\newblock
{\BBOQ}\APACrefatitle {Data Statements for Natural Language Processing: Toward
  Mitigating System Bias and Enabling Better Science} {Data statements for
  natural language processing: Toward mitigating system bias and enabling
  better science}.{\BBCQ}
\newblock
\APACjournalVolNumPages{Transactions of the Association for Computational
  Linguistics}{6}{}{587--604,}
\newblock
\begin{APACrefDOI} \doi{10.1162/tacl_a_00041} \end{APACrefDOI}
\newblock

\newblock

\PrintBackRefs{\CurrentBib}

\bibitem [\protect \citeauthoryear {%
Blodgett%
, Barocas%
, Daum{\'e}~III%
\BCBL {}\ \BBA {} Wallach%
}{%
Blodgett%
\ \protect \BOthers {.}}{%
{\protect \APACyear {2020}}%
}]{%
blodgett2020power}
\APACinsertmetastar {%
blodgett2020power}%
\begin{APACrefauthors}%
Blodgett, S.L.%
, Barocas, S.%
, Daum{\'e}~III, H.%
\BCBL {} Wallach, H.%
\end{APACrefauthors}%
\unskip\
\newblock
\APACrefYearMonthDay{2020}{}{}.
\newblock
{\BBOQ}\APACrefatitle {Language (Technology) is Power: A Critical Survey of
  ``Bias'' in {NLP}} {Language (technology) is power: A critical survey of
  ``bias'' in {NLP}}.{\BBCQ}
\newblock
 \APACrefbtitle {{Proceedings of the 58th Annual Meeting of the Association for
  Computational Linguistics}} {{Proceedings of the 58th Annual Meeting of the
  Association for Computational Linguistics}}\ (\BPGS\ 5454--5476).
\PrintBackRefs{\CurrentBib}

\bibitem [\protect \citeauthoryear {%
Bollmann%
}{%
Bollmann%
}{%
{\protect \APACyear {2019}}%
}]{%
bollmann2019normalization}
\APACinsertmetastar {%
bollmann2019normalization}%
\begin{APACrefauthors}%
Bollmann, M.%
\end{APACrefauthors}%
\unskip\
\newblock
\APACrefYearMonthDay{2019}{}{}.
\newblock
{\BBOQ}\APACrefatitle {A Large-Scale Comparison of Historical Text
  Normalization Systems} {A large-scale comparison of historical text
  normalization systems}.{\BBCQ}
\newblock
 \APACrefbtitle {{Proceedings of the 2019 Conference of the North American
  Chapter of the Association for Computational Linguistics: Human Language
  Technologies}} {{Proceedings of the 2019 Conference of the North American
  Chapter of the Association for Computational Linguistics: Human Language
  Technologies}}\ (\BPGS\ 3885--3898).
\PrintBackRefs{\CurrentBib}

\bibitem [\protect \citeauthoryear {%
Broder%
}{%
Broder%
}{%
{\protect \APACyear {1997}}%
}]{%
broder1997minhash}
\APACinsertmetastar {%
broder1997minhash}%
\begin{APACrefauthors}%
Broder, A.Z.%
\end{APACrefauthors}%
\unskip\
\newblock
\APACrefYearMonthDay{1997}{}{}.
\newblock
{\BBOQ}\APACrefatitle {On the Resemblance and Containment of Documents} {On the
  resemblance and containment of documents}.{\BBCQ}
\newblock
 \APACrefbtitle {{Proceedings of the Compression and Complexity of Sequences}}
  {{Proceedings of the Compression and Complexity of Sequences}}\ (\BPGS\
  21--29).
\newblock
\APACaddressPublisher{}{IEEE}.
\PrintBackRefs{\CurrentBib}

\bibitem [\protect \citeauthoryear {%
Byszuk%
}{%
Byszuk%
}{%
{\protect \APACyear {2021}}%
}]{%
byszuk2021eltecpol}
\APACinsertmetastar {%
byszuk2021eltecpol}%
\begin{APACrefauthors}%
Byszuk, J.%
\end{APACrefauthors}%
\unskip\
\newblock
\APACrefYearMonthDay{2021}{}{}.
\newblock
\APACrefbtitle {Polish Novel Collection ({ELTeC-pol}).} {Polish novel
  collection ({ELTeC-pol}).}
\newblock
\APAChowpublished {Edited collection, version 1.0.0. In: European Literary Text
  Collection (ELTeC), COST Action Distant Reading for European Literary
  History}.
\PrintBackRefs{\CurrentBib}

\bibitem [\protect \citeauthoryear {%
Dao%
, Fu%
, Ermon%
, Rudra%
\BCBL {}\ \BBA {} R{\'e}%
}{%
Dao%
\ \protect \BOthers {.}}{%
{\protect \APACyear {2022}}%
}]{%
dao2022flashattention}
\APACinsertmetastar {%
dao2022flashattention}%
\begin{APACrefauthors}%
Dao, T.%
, Fu, D.Y.%
, Ermon, S.%
, Rudra, A.%
\BCBL {} R{\'e}, C.%
\end{APACrefauthors}%
\unskip\
\newblock
\APACrefYearMonthDay{2022}{}{}.
\newblock
{\BBOQ}\APACrefatitle {{FlashAttention}: Fast and Memory-Efficient Exact
  Attention with {IO}-Awareness} {{FlashAttention}: Fast and memory-efficient
  exact attention with {IO}-awareness}.{\BBCQ}
\newblock
 \APACrefbtitle {{Advances in Neural Information Processing Systems
  (NeurIPS)}.} {{Advances in Neural Information Processing Systems (NeurIPS)}.}
\PrintBackRefs{\CurrentBib}

\bibitem [\protect \citeauthoryear {%
{DeepSeek-AI}%
}{%
{DeepSeek-AI}%
}{%
{\protect \APACyear {2026}}%
}]{%
deepseek2026v4}
\APACinsertmetastar {%
deepseek2026v4}%
\begin{APACrefauthors}%
{DeepSeek-AI}%
\end{APACrefauthors}%
\unskip\
\newblock
\APACrefYearMonthDay{2026}{}{}.
\newblock
\APACrefbtitle {{DeepSeek-V4}: Towards Highly Efficient Million-Token Context
  Intelligence.} {{DeepSeek-V4}: Towards highly efficient million-token context
  intelligence.}
\newblock
\begin{APACrefURL} {https://arxiv.org/abs/2606.19348} \end{APACrefURL}
\PrintBackRefs{\CurrentBib}

\bibitem [\protect \citeauthoryear {%
Dhingra%
\ \protect \BOthers {.}}{%
Dhingra%
\ \protect \BOthers {.}}{%
{\protect \APACyear {2022}}%
}]{%
dhingra2022timeaware}
\APACinsertmetastar {%
dhingra2022timeaware}%
\begin{APACrefauthors}%
Dhingra, B.%
, Cole, J.R.%
, Eisenschlos, J.M.%
, Gillick, D.%
, Eisenstein, J.%
\BCBL {} Cohen, W.W.%
\end{APACrefauthors}%
\unskip\
\newblock
\APACrefYearMonthDay{2022}{}{}.
\newblock
{\BBOQ}\APACrefatitle {Time-Aware Language Models as Temporal Knowledge Bases}
  {Time-aware language models as temporal knowledge bases}.{\BBCQ}
\newblock
\APACjournalVolNumPages{Transactions of the Association for Computational
  Linguistics}{10}{}{257--273,}
\newblock

\newblock

\PrintBackRefs{\CurrentBib}

\bibitem [\protect \citeauthoryear {%
Dodge%
\ \protect \BOthers {.}}{%
Dodge%
\ \protect \BOthers {.}}{%
{\protect \APACyear {2021}}%
}]{%
dodge2021documenting}
\APACinsertmetastar {%
dodge2021documenting}%
\begin{APACrefauthors}%
Dodge, J.%
, Sap, M.%
, Marasovi{\'c}, A.%
, Agnew, W.%
, Ilharco, G.%
, Groeneveld, D.%
\BDBL {}Gardner, M.%
\end{APACrefauthors}%
\unskip\
\newblock
\APACrefYearMonthDay{2021}{}{}.
\newblock
{\BBOQ}\APACrefatitle {Documenting Large Webtext Corpora: A Case Study on the
  Colossal Clean Crawled Corpus} {Documenting large webtext corpora: A case
  study on the colossal clean crawled corpus}.{\BBCQ}
\newblock
 \APACrefbtitle {{Proceedings of the 2021 Conference on Empirical Methods in
  Natural Language Processing}} {{Proceedings of the 2021 Conference on
  Empirical Methods in Natural Language Processing}}\ (\BPGS\ 1286--1305).
\PrintBackRefs{\CurrentBib}

\bibitem [\protect \citeauthoryear {%
Drinkall%
, Rahimikia%
, Pierrehumbert%
\BCBL {}\ \BBA {} Zohren%
}{%
Drinkall%
\ \protect \BOthers {.}}{%
{\protect \APACyear {2024}}%
}]{%
drinkall2024timemachine}
\APACinsertmetastar {%
drinkall2024timemachine}%
\begin{APACrefauthors}%
Drinkall, F.%
, Rahimikia, E.%
, Pierrehumbert, J.%
\BCBL {} Zohren, S.%
\end{APACrefauthors}%
\unskip\
\newblock
\APACrefYearMonthDay{2024}{}{}.
\newblock
{\BBOQ}\APACrefatitle {Time Machine {GPT}} {Time machine {GPT}}.{\BBCQ}
\newblock
 \APACrefbtitle {{Findings of the Association for Computational Linguistics:
  NAACL 2024}} {{Findings of the Association for Computational Linguistics:
  NAACL 2024}}\ (\BPGS\ 3281--3292).
\PrintBackRefs{\CurrentBib}

\bibitem [\protect \citeauthoryear {%
Elazar%
\ \protect \BOthers {.}}{%
Elazar%
\ \protect \BOthers {.}}{%
{\protect \APACyear {2024}}%
}]{%
elazar2024wimbd}
\APACinsertmetastar {%
elazar2024wimbd}%
\begin{APACrefauthors}%
Elazar, Y.%
, Bhagia, A.%
, Magnusson, I.%
, Ravichander, A.%
, Schwenk, D.%
, Suhr, A.%
\BDBL {}Dodge, J.%
\end{APACrefauthors}%
\unskip\
\newblock
\APACrefYearMonthDay{2024}{}{}.
\newblock
{\BBOQ}\APACrefatitle {What's In My Big Data?} {What's in my big data?}{\BBCQ}
\newblock
 \APACrefbtitle {{International Conference on Learning Representations
  (ICLR)}.} {{International Conference on Learning Representations (ICLR)}.}
\PrintBackRefs{\CurrentBib}

\bibitem [\protect \citeauthoryear {%
Gao%
\ \protect \BOthers {.}}{%
Gao%
\ \protect \BOthers {.}}{%
{\protect \APACyear {2020}}%
}]{%
gao2020pile}
\APACinsertmetastar {%
gao2020pile}%
\begin{APACrefauthors}%
Gao, L.%
, Biderman, S.%
, Black, S.%
, Golding, L.%
, Hoppe, T.%
, Foster, C.%
\BDBL {}Leahy, C.%
\end{APACrefauthors}%
\unskip\
\newblock
\APACrefYearMonthDay{2020}{}{}.
\newblock
\APACrefbtitle {The {P}ile: An 800{GB} Dataset of Diverse Text for Language
  Modeling.} {The {P}ile: An 800{GB} dataset of diverse text for language
  modeling.}
\newblock
\begin{APACrefURL} {https://arxiv.org/abs/2101.00027} \end{APACrefURL}
\PrintBackRefs{\CurrentBib}

\bibitem [\protect \citeauthoryear {%
Gebru%
\ \protect \BOthers {.}}{%
Gebru%
\ \protect \BOthers {.}}{%
{\protect \APACyear {2021}}%
}]{%
gebru2021datasheets}
\APACinsertmetastar {%
gebru2021datasheets}%
\begin{APACrefauthors}%
Gebru, T.%
, Morgenstern, J.%
, Vecchione, B.%
, Wortman~Vaughan, J.%
, Wallach, H.%
, Daum{\'e}~III, H.%
\BCBL {} Crawford, K.%
\end{APACrefauthors}%
\unskip\
\newblock
\APACrefYearMonthDay{2021}{}{}.
\newblock
{\BBOQ}\APACrefatitle {Datasheets for Datasets} {Datasheets for
  datasets}.{\BBCQ}
\newblock
\APACjournalVolNumPages{Communications of the ACM}{64}{12}{86--92,}
\newblock
\begin{APACrefDOI} \doi{10.1145/3458723} \end{APACrefDOI}
\newblock

\newblock

\PrintBackRefs{\CurrentBib}

\bibitem [\protect \citeauthoryear {%
G{\"o}ttlich%
, Loibner%
, Jiang%
\BCBL {}\ \BBA {} Voth%
}{%
G{\"o}ttlich%
\ \protect \BOthers {.}}{%
{\protect \APACyear {2025}}%
}]{%
goettlich2025historyllms}
\APACinsertmetastar {%
goettlich2025historyllms}%
\begin{APACrefauthors}%
G{\"o}ttlich, D.%
, Loibner, D.%
, Jiang, G.%
\BCBL {} Voth, H\BHBI J.%
\end{APACrefauthors}%
\unskip\
\newblock
\APACrefYearMonthDay{2025}{}{}.
\newblock
\APACrefbtitle {History {LLMs}} {History {LLMs}}\ \APACbVolEdTR{}{\BTR{}}.
\newblock
\APACaddressInstitution{}{University of Zurich and Cologne University}.
\newblock
\begin{APACrefURL} {https://github.com/DGoettlich/history-llms}
  \end{APACrefURL}
\PrintBackRefs{\CurrentBib}

\bibitem [\protect \citeauthoryear {%
Grigorian%
\ \BBA {} Yaghoobian%
}{%
Grigorian%
\ \BBA {} Yaghoobian%
}{%
{\protect \APACyear {2025}}%
}]{%
timecapsulellm}
\APACinsertmetastar {%
timecapsulellm}%
\begin{APACrefauthors}%
Grigorian, H.%
\BCBT {}\ \BBA {} Yaghoobian, H.%
\end{APACrefauthors}%
\unskip\
\newblock
\APACrefYearMonthDay{2025}{}{}.
\newblock
\APACrefbtitle {{TimeCapsuleLLM}: A {LLM} Trained Only on Data from Certain
  Time Periods to Reduce Modern Bias.} {{TimeCapsuleLLM}: A {LLM} trained only
  on data from certain time periods to reduce modern bias.}
\newblock
\APAChowpublished {Software repository,
  \url{https://github.com/haykgrigo3/TimeCapsuleLLM}}.
\PrintBackRefs{\CurrentBib}

\bibitem [\protect \citeauthoryear {%
Grigorian%
\ \BBA {} Yaghoobian%
}{%
Grigorian%
\ \BBA {} Yaghoobian%
}{%
{\protect \APACyear {2026}}%
}]{%
grigorian2026timecapsule}
\APACinsertmetastar {%
grigorian2026timecapsule}%
\begin{APACrefauthors}%
Grigorian, H.%
\BCBT {}\ \BBA {} Yaghoobian, H.%
\end{APACrefauthors}%
\unskip\
\newblock
\APACrefYearMonthDay{2026}{}{}.
\newblock
{\BBOQ}\APACrefatitle {{TimeCapsule}: Generative Hallucination as a Method for
  Historical Sensemaking} {{TimeCapsule}: Generative hallucination as a method
  for historical sensemaking}.{\BBCQ}
\newblock
 \APACrefbtitle {{Proceedings of the 2026 Conference on Creativity and
  Cognition}} {{Proceedings of the 2026 Conference on Creativity and
  Cognition}}\ (\BPGS\ 229--238).
\newblock
\APACaddressPublisher{London, United Kingdom}{ACM}.
\PrintBackRefs{\CurrentBib}

\bibitem [\protect \citeauthoryear {%
Gruszczy{\'n}ski%
\ \protect \BOthers {.}}{%
Gruszczy{\'n}ski%
\ \protect \BOthers {.}}{%
{\protect \APACyear {2022}}%
}]{%
gruszczynski2022korba}
\APACinsertmetastar {%
gruszczynski2022korba}%
\begin{APACrefauthors}%
Gruszczy{\'n}ski, W.%
, Adamiec, D.%
, Bronikowska, R.%
, Kiera{\'s}, W.%
, Modrzejewski, E.%
, Wieczorek, A.%
\BCBL {} Woli{\'n}ski, M.%
\end{APACrefauthors}%
\unskip\
\newblock
\APACrefYearMonthDay{2022}{}{}.
\newblock
{\BBOQ}\APACrefatitle {The Electronic Corpus of 17th- and 18th-century {P}olish
  Texts} {The electronic corpus of 17th- and 18th-century {P}olish
  texts}.{\BBCQ}
\newblock
\APACjournalVolNumPages{Language Resources and Evaluation}{56}{1}{309--332,}
\newblock
\begin{APACrefDOI} \doi{10.1007/s10579-021-09549-1} \end{APACrefDOI}
\newblock

\newblock

\PrintBackRefs{\CurrentBib}

\bibitem [\protect \citeauthoryear {%
H{\"a}gele%
\ \protect \BOthers {.}}{%
H{\"a}gele%
\ \protect \BOthers {.}}{%
{\protect \APACyear {2024}}%
}]{%
hagele2024scaling}
\APACinsertmetastar {%
hagele2024scaling}%
\begin{APACrefauthors}%
H{\"a}gele, A.%
, Bakouch, E.%
, Kosson, A.%
, Ben~Allal, L.%
, von Werra, L.%
\BCBL {} Jaggi, M.%
\end{APACrefauthors}%
\unskip\
\newblock
\APACrefYearMonthDay{2024}{}{}.
\newblock
{\BBOQ}\APACrefatitle {Scaling Laws and Compute-Optimal Training Beyond Fixed
  Training Durations} {Scaling laws and compute-optimal training beyond fixed
  training durations}.{\BBCQ}
\newblock
 \APACrefbtitle {{Advances in Neural Information Processing Systems
  (NeurIPS)}.} {{Advances in Neural Information Processing Systems (NeurIPS)}.}
\PrintBackRefs{\CurrentBib}

\bibitem [\protect \citeauthoryear {%
He%
, Lv%
, Manela%
\BCBL {}\ \BBA {} Wu%
}{%
He%
\ \protect \BOthers {.}}{%
{\protect \APACyear {2025}}%
}]{%
he2025chronologically}
\APACinsertmetastar {%
he2025chronologically}%
\begin{APACrefauthors}%
He, S.%
, Lv, L.%
, Manela, A.%
\BCBL {} Wu, J.%
\end{APACrefauthors}%
\unskip\
\newblock
\APACrefYearMonthDay{2025}{}{}.
\newblock
\APACrefbtitle {Chronologically Consistent Large Language Models.}
  {Chronologically consistent large language models.}
\newblock
\begin{APACrefURL} {https://arxiv.org/abs/2502.21206} \end{APACrefURL}
\PrintBackRefs{\CurrentBib}

\bibitem [\protect \citeauthoryear {%
Hill%
\ \BBA {} Hengchen%
}{%
Hill%
\ \BBA {} Hengchen%
}{%
{\protect \APACyear {2019}}%
}]{%
hill2019dirtyocr}
\APACinsertmetastar {%
hill2019dirtyocr}%
\begin{APACrefauthors}%
Hill, M.J.%
\BCBT {}\ \BBA {} Hengchen, S.%
\end{APACrefauthors}%
\unskip\
\newblock
\APACrefYearMonthDay{2019}{}{}.
\newblock
{\BBOQ}\APACrefatitle {Quantifying the Impact of Dirty {OCR} on Historical Text
  Analysis: {E}ighteenth {C}entury {C}ollections {O}nline as a Case Study}
  {Quantifying the impact of dirty {OCR} on historical text analysis:
  {E}ighteenth {C}entury {C}ollections {O}nline as a case study}.{\BBCQ}
\newblock
\APACjournalVolNumPages{Digital Scholarship in the
  Humanities}{34}{4}{825--843,}
\newblock
\begin{APACrefDOI} \doi{10.1093/llc/fqz024} \end{APACrefDOI}
\newblock

\newblock

\PrintBackRefs{\CurrentBib}

\bibitem [\protect \citeauthoryear {%
Hoffmann%
\ \protect \BOthers {.}}{%
Hoffmann%
\ \protect \BOthers {.}}{%
{\protect \APACyear {2022}}%
}]{%
hoffmann2022chinchilla}
\APACinsertmetastar {%
hoffmann2022chinchilla}%
\begin{APACrefauthors}%
Hoffmann, J.%
, Borgeaud, S.%
, Mensch, A.%
, Buchatskaya, E.%
, Cai, T.%
, Rutherford, E.%
\BDBL {}Sifre, L.%
\end{APACrefauthors}%
\unskip\
\newblock
\APACrefYearMonthDay{2022}{}{}.
\newblock
{\BBOQ}\APACrefatitle {An Empirical Analysis of Compute-Optimal Large Language
  Model Training} {An empirical analysis of compute-optimal large language
  model training}.{\BBCQ}
\newblock
 \APACrefbtitle {{Advances in Neural Information Processing Systems
  (NeurIPS)}.} {{Advances in Neural Information Processing Systems (NeurIPS)}.}
\PrintBackRefs{\CurrentBib}

\bibitem [\protect \citeauthoryear {%
Jordan%
\ \protect \BOthers {.}}{%
Jordan%
\ \protect \BOthers {.}}{%
{\protect \APACyear {2024}}%
}]{%
jordan2024muon}
\APACinsertmetastar {%
jordan2024muon}%
\begin{APACrefauthors}%
Jordan, K.%
, Jin, Y.%
, Boza, V.%
, You, J.%
, Cesista, F.%
, Newhouse, L.%
\BCBL {} Bernstein, J.%
\end{APACrefauthors}%
\unskip\
\newblock
\APACrefYearMonthDay{2024}{}{}.
\newblock
\APACrefbtitle {Muon: An Optimizer for Hidden Layers in Neural Networks.}
  {Muon: An optimizer for hidden layers in neural networks.}
\newblock
\APAChowpublished {\url{https://kellerjordan.github.io/posts/muon/}}.
\PrintBackRefs{\CurrentBib}

\bibitem [\protect \citeauthoryear {%
Kaplan%
\ \protect \BOthers {.}}{%
Kaplan%
\ \protect \BOthers {.}}{%
{\protect \APACyear {2020}}%
}]{%
kaplan2020scaling}
\APACinsertmetastar {%
kaplan2020scaling}%
\begin{APACrefauthors}%
Kaplan, J.%
, McCandlish, S.%
, Henighan, T.%
, Brown, T.B.%
, Chess, B.%
, Child, R.%
\BDBL {}Amodei, D.%
\end{APACrefauthors}%
\unskip\
\newblock
\APACrefYearMonthDay{2020}{}{}.
\newblock
\APACrefbtitle {Scaling Laws for Neural Language Models.} {Scaling laws for
  neural language models.}
\newblock
\begin{APACrefURL} {https://arxiv.org/abs/2001.08361} \end{APACrefURL}
\PrintBackRefs{\CurrentBib}

\bibitem [\protect \citeauthoryear {%
Kiera{\'s}%
\ \BBA {} Woli{\'n}ski%
}{%
Kiera{\'s}%
\ \BBA {} Woli{\'n}ski%
}{%
{\protect \APACyear {2018}}%
}]{%
kieras2018f19}
\APACinsertmetastar {%
kieras2018f19}%
\begin{APACrefauthors}%
Kiera{\'s}, W.%
\BCBT {}\ \BBA {} Woli{\'n}ski, M.%
\end{APACrefauthors}%
\unskip\
\newblock
\APACrefYearMonthDay{2018}{}{}.
\newblock
{\BBOQ}\APACrefatitle {Manually Annotated Corpus of {P}olish Texts Published
  between 1830 and 1918} {Manually annotated corpus of {P}olish texts published
  between 1830 and 1918}.{\BBCQ}
\newblock
 \APACrefbtitle {{Proceedings of the Eleventh International Conference on
  Language Resources and Evaluation (LREC 2018)}.} {{Proceedings of the
  Eleventh International Conference on Language Resources and Evaluation (LREC
  2018)}.}
\newblock
\begin{APACrefURL} {https://aclanthology.org/L18-1609/} \end{APACrefURL}
\PrintBackRefs{\CurrentBib}

\bibitem [\protect \citeauthoryear {%
Koco{\'n}%
, Piasecki%
, Janz%
\BCBL {}\ \protect \BOthers {.}}{%
Koco{\'n}%
\ \protect \BOthers {.}}{%
{\protect \APACyear {2025}}%
}]{%
kocon2025pllum}
\APACinsertmetastar {%
kocon2025pllum}%
\begin{APACrefauthors}%
Koco{\'n}, J.%
, Piasecki, M.%
, Janz, A.%
\BCBL {}\ \BOthersPeriod {.}\end{APACrefauthors}%
\unskip\
\newblock
\APACrefYearMonthDay{2025}{}{}.
\newblock
\APACrefbtitle {{PLLuM}: A Family of {P}olish Large Language Models.} {{PLLuM}:
  A family of {P}olish large language models.}
\newblock
\begin{APACrefURL} {https://arxiv.org/abs/2511.03823} \end{APACrefURL}
\PrintBackRefs{\CurrentBib}

\bibitem [\protect \citeauthoryear {%
Lauren{\c{c}}on%
\ \protect \BOthers {.}}{%
Lauren{\c{c}}on%
\ \protect \BOthers {.}}{%
{\protect \APACyear {2022}}%
}]{%
laurencon2022roots}
\APACinsertmetastar {%
laurencon2022roots}%
\begin{APACrefauthors}%
Lauren{\c{c}}on, H.%
, Saulnier, L.%
, Wang, T.%
, Akiki, C.%
, Villanova~del Moral, A.%
, Le~Scao, T.%
\BDBL {}Jernite, Y.%
\end{APACrefauthors}%
\unskip\
\newblock
\APACrefYearMonthDay{2022}{}{}.
\newblock
{\BBOQ}\APACrefatitle {The {B}ig{S}cience {ROOTS} Corpus: A 1.6{TB} Composite
  Multilingual Dataset} {The {B}ig{S}cience {ROOTS} corpus: A 1.6{TB} composite
  multilingual dataset}.{\BBCQ}
\newblock
 \APACrefbtitle {{Advances in Neural Information Processing Systems (NeurIPS),
  Datasets and Benchmarks Track}.} {{Advances in Neural Information Processing
  Systems (NeurIPS), Datasets and Benchmarks Track}.}
\PrintBackRefs{\CurrentBib}

\bibitem [\protect \citeauthoryear {%
Lazaridou%
\ \protect \BOthers {.}}{%
Lazaridou%
\ \protect \BOthers {.}}{%
{\protect \APACyear {2021}}%
}]{%
lazaridou2021mindthegap}
\APACinsertmetastar {%
lazaridou2021mindthegap}%
\begin{APACrefauthors}%
Lazaridou, A.%
, Kuncoro, A.%
, Gribovskaya, E.%
, Agrawal, D.%
, Liska, A.%
, Terzi, T.%
\BDBL {}Blunsom, P.%
\end{APACrefauthors}%
\unskip\
\newblock
\APACrefYearMonthDay{2021}{}{}.
\newblock
{\BBOQ}\APACrefatitle {Mind the Gap: Assessing Temporal Generalization in
  Neural Language Models} {Mind the gap: Assessing temporal generalization in
  neural language models}.{\BBCQ}
\newblock
 \APACrefbtitle {{Advances in Neural Information Processing Systems 34 (NeurIPS
  2021)}.} {{Advances in Neural Information Processing Systems 34 (NeurIPS
  2021)}.}
\PrintBackRefs{\CurrentBib}

\bibitem [\protect \citeauthoryear {%
Lee%
\ \protect \BOthers {.}}{%
Lee%
\ \protect \BOthers {.}}{%
{\protect \APACyear {2022}}%
}]{%
lee2022dedup}
\APACinsertmetastar {%
lee2022dedup}%
\begin{APACrefauthors}%
Lee, K.%
, Ippolito, D.%
, Nystrom, A.%
, Zhang, C.%
, Eck, D.%
, Callison-Burch, C.%
\BCBL {} Carlini, N.%
\end{APACrefauthors}%
\unskip\
\newblock
\APACrefYearMonthDay{2022}{}{}.
\newblock
{\BBOQ}\APACrefatitle {Deduplicating Training Data Makes Language Models
  Better} {Deduplicating training data makes language models better}.{\BBCQ}
\newblock
 \APACrefbtitle {{Proceedings of the 60th Annual Meeting of the Association for
  Computational Linguistics (ACL)}} {{Proceedings of the 60th Annual Meeting of
  the Association for Computational Linguistics (ACL)}}\ (\BPGS\ 8424--8445).
\newblock
\begin{APACrefURL} {https://aclanthology.org/2022.acl-long.577/}
  \end{APACrefURL}
\PrintBackRefs{\CurrentBib}

\bibitem [\protect \citeauthoryear {%
Loshchilov%
\ \BBA {} Hutter%
}{%
Loshchilov%
\ \BBA {} Hutter%
}{%
{\protect \APACyear {2019}}%
}]{%
loshchilov2019adamw}
\APACinsertmetastar {%
loshchilov2019adamw}%
\begin{APACrefauthors}%
Loshchilov, I.%
\BCBT {}\ \BBA {} Hutter, F.%
\end{APACrefauthors}%
\unskip\
\newblock
\APACrefYearMonthDay{2019}{}{}.
\newblock
{\BBOQ}\APACrefatitle {Decoupled Weight Decay Regularization} {Decoupled weight
  decay regularization}.{\BBCQ}
\newblock
 \APACrefbtitle {{International Conference on Learning Representations
  (ICLR)}.} {{International Conference on Learning Representations (ICLR)}.}
\PrintBackRefs{\CurrentBib}

\bibitem [\protect \citeauthoryear {%
Luo%
\ \protect \BOthers {.}}{%
Luo%
\ \protect \BOthers {.}}{%
{\protect \APACyear {2026}}%
}]{%
luo2026typewriterlm}
\APACinsertmetastar {%
luo2026typewriterlm}%
\begin{APACrefauthors}%
Luo, X.%
, Shinnick, Z.%
, Griesshaber, N.%
, Wang, Y.%
, Yu, J.%
, Shi, F.%
\BDBL {}Lu, Y.%
\end{APACrefauthors}%
\unskip\
\newblock
\APACrefYearMonthDay{2026}{}{}.
\newblock
\APACrefbtitle {A Language Model from 1913: Pretraining on Historical Text.} {A
  language model from 1913: Pretraining on historical text.}
\newblock
\begin{APACrefURL} {https://arxiv.org/abs/2606.02991} \end{APACrefURL}
\PrintBackRefs{\CurrentBib}

\bibitem [\protect \citeauthoryear {%
Luu%
, Khashabi%
, Gururangan%
, Mandyam%
\BCBL {}\ \BBA {} Smith%
}{%
Luu%
\ \protect \BOthers {.}}{%
{\protect \APACyear {2022}}%
}]{%
luu2022timewaits}
\APACinsertmetastar {%
luu2022timewaits}%
\begin{APACrefauthors}%
Luu, K.%
, Khashabi, D.%
, Gururangan, S.%
, Mandyam, K.%
\BCBL {} Smith, N.A.%
\end{APACrefauthors}%
\unskip\
\newblock
\APACrefYearMonthDay{2022}{}{}.
\newblock
{\BBOQ}\APACrefatitle {Time Waits for No One! {A}nalysis and Challenges of
  Temporal Misalignment} {Time waits for no one! {A}nalysis and challenges of
  temporal misalignment}.{\BBCQ}
\newblock
 \APACrefbtitle {{Proceedings of the 2022 Conference of the North American
  Chapter of the Association for Computational Linguistics: Human Language
  Technologies}} {{Proceedings of the 2022 Conference of the North American
  Chapter of the Association for Computational Linguistics: Human Language
  Technologies}}\ (\BPGS\ 5944--5958).
\PrintBackRefs{\CurrentBib}

\bibitem [\protect \citeauthoryear {%
{Mistral AI}%
}{%
{Mistral AI}%
}{%
{\protect \APACyear {2025}}%
}]{%
mistral2025large3}
\APACinsertmetastar {%
mistral2025large3}%
\begin{APACrefauthors}%
{Mistral AI}%
\end{APACrefauthors}%
\unskip\
\newblock
\APACrefYearMonthDay{2025}{}{}.
\newblock
\APACrefbtitle {{Mistral-Large-3-675B-Instruct-2512}.}
  {{Mistral-Large-3-675B-Instruct-2512}.}
\newblock
\APAChowpublished {Model repository,
  \url{https://huggingface.co/mistralai/Mistral-Large-3-675B-Instruct-2512}}.
\PrintBackRefs{\CurrentBib}

\bibitem [\protect \citeauthoryear {%
Mitchell%
\ \protect \BOthers {.}}{%
Mitchell%
\ \protect \BOthers {.}}{%
{\protect \APACyear {2019}}%
}]{%
mitchell2019modelcards}
\APACinsertmetastar {%
mitchell2019modelcards}%
\begin{APACrefauthors}%
Mitchell, M.%
, Wu, S.%
, Zaldivar, A.%
, Barnes, P.%
, Vasserman, L.%
, Hutchinson, B.%
\BDBL {}Gebru, T.%
\end{APACrefauthors}%
\unskip\
\newblock
\APACrefYearMonthDay{2019}{}{}.
\newblock
{\BBOQ}\APACrefatitle {Model Cards for Model Reporting} {Model cards for model
  reporting}.{\BBCQ}
\newblock
 \APACrefbtitle {{Proceedings of the Conference on Fairness, Accountability,
  and Transparency (FAT*)}} {{Proceedings of the Conference on Fairness,
  Accountability, and Transparency (FAT*)}}\ (\BPGS\ 220--229).
\newblock
\APACaddressPublisher{}{ACM}.
\PrintBackRefs{\CurrentBib}

\bibitem [\protect \citeauthoryear {%
Mroczkowski%
, Rybak%
, Wr{\'o}blewska%
\BCBL {}\ \BBA {} Gawlik%
}{%
Mroczkowski%
\ \protect \BOthers {.}}{%
{\protect \APACyear {2021}}%
}]{%
mroczkowski2021herbert}
\APACinsertmetastar {%
mroczkowski2021herbert}%
\begin{APACrefauthors}%
Mroczkowski, R.%
, Rybak, P.%
, Wr{\'o}blewska, A.%
\BCBL {} Gawlik, I.%
\end{APACrefauthors}%
\unskip\
\newblock
\APACrefYearMonthDay{2021}{}{}.
\newblock
{\BBOQ}\APACrefatitle {{HerBERT}: Efficiently Pretrained Transformer-based
  Language Model for {P}olish} {{HerBERT}: Efficiently pretrained
  transformer-based language model for {P}olish}.{\BBCQ}
\newblock
 \APACrefbtitle {{Proceedings of the 8th Workshop on Balto-Slavic Natural
  Language Processing}.} {{Proceedings of the 8th Workshop on Balto-Slavic
  Natural Language Processing}.}
\newblock
\begin{APACrefURL} {https://aclanthology.org/2021.bsnlp-1.1/} \end{APACrefURL}
\PrintBackRefs{\CurrentBib}

\bibitem [\protect \citeauthoryear {%
Muennighoff%
\ \protect \BOthers {.}}{%
Muennighoff%
\ \protect \BOthers {.}}{%
{\protect \APACyear {2023}}%
}]{%
muennighoff2023scaling}
\APACinsertmetastar {%
muennighoff2023scaling}%
\begin{APACrefauthors}%
Muennighoff, N.%
, Rush, A.M.%
, Barak, B.%
, Le~Scao, T.%
, Piktus, A.%
, Tazi, N.%
\BDBL {}Raffel, C.%
\end{APACrefauthors}%
\unskip\
\newblock
\APACrefYearMonthDay{2023}{}{}.
\newblock
{\BBOQ}\APACrefatitle {Scaling Data-Constrained Language Models} {Scaling
  data-constrained language models}.{\BBCQ}
\newblock
 \APACrefbtitle {{Advances in Neural Information Processing Systems
  (NeurIPS)}.} {{Advances in Neural Information Processing Systems (NeurIPS)}.}
\PrintBackRefs{\CurrentBib}

\bibitem [\protect \citeauthoryear {%
Nangia%
, Vania%
, Bhalerao%
\BCBL {}\ \BBA {} Bowman%
}{%
Nangia%
\ \protect \BOthers {.}}{%
{\protect \APACyear {2020}}%
}]{%
nangia2020crowspairs}
\APACinsertmetastar {%
nangia2020crowspairs}%
\begin{APACrefauthors}%
Nangia, N.%
, Vania, C.%
, Bhalerao, R.%
\BCBL {} Bowman, S.R.%
\end{APACrefauthors}%
\unskip\
\newblock
\APACrefYearMonthDay{2020}{}{}.
\newblock
{\BBOQ}\APACrefatitle {{C}row{S}-Pairs: A Challenge Dataset for Measuring
  Social Biases in Masked Language Models} {{C}row{S}-pairs: A challenge
  dataset for measuring social biases in masked language models}.{\BBCQ}
\newblock
 \APACrefbtitle {{Proceedings of the 2020 Conference on Empirical Methods in
  Natural Language Processing (EMNLP)}} {{Proceedings of the 2020 Conference on
  Empirical Methods in Natural Language Processing (EMNLP)}}\ (\BPGS\
  1953--1967).
\PrintBackRefs{\CurrentBib}

\bibitem [\protect \citeauthoryear {%
Nguyen%
\ \protect \BOthers {.}}{%
Nguyen%
\ \protect \BOthers {.}}{%
{\protect \APACyear {2024}}%
}]{%
nguyen2024culturax}
\APACinsertmetastar {%
nguyen2024culturax}%
\begin{APACrefauthors}%
Nguyen, T.%
, Nguyen, C.V.%
, Lai, V.D.%
, Man, H.%
, Ngo, N.T.%
, Dernoncourt, F.%
\BDBL {}Nguyen, T.H.%
\end{APACrefauthors}%
\unskip\
\newblock
\APACrefYearMonthDay{2024}{}{}.
\newblock
{\BBOQ}\APACrefatitle {{C}ultura{X}: A Cleaned, Enormous, and Multilingual
  Dataset for Large Language Models in 167 Languages} {{C}ultura{X}: A cleaned,
  enormous, and multilingual dataset for large language models in 167
  languages}.{\BBCQ}
\newblock
 \APACrefbtitle {{Proceedings of the 2024 Joint International Conference on
  Computational Linguistics, Language Resources and Evaluation (LREC-COLING
  2024)}} {{Proceedings of the 2024 Joint International Conference on
  Computational Linguistics, Language Resources and Evaluation (LREC-COLING
  2024)}}\ (\BPGS\ 4226--4237).
\newblock
\APACaddressPublisher{}{ELRA and ICCL}.
\newblock
\begin{APACrefURL} {https://aclanthology.org/2024.lrec-main.377/}
  \end{APACrefURL}
\PrintBackRefs{\CurrentBib}

\bibitem [\protect \citeauthoryear {%
Ociepa%
, Flis%
, Kinas%
, Wr{\'o}bel%
\BCBL {}\ \BBA {} Gwo{\'z}dziej%
}{%
Ociepa%
\ \protect \BOthers {.}}{%
{\protect \APACyear {2025}}%
}]{%
ociepa2025bielikv3}
\APACinsertmetastar {%
ociepa2025bielikv3}%
\begin{APACrefauthors}%
Ociepa, K.%
, Flis, {\L}.%
, Kinas, R.%
, Wr{\'o}bel, K.%
\BCBL {} Gwo{\'z}dziej, A.%
\end{APACrefauthors}%
\unskip\
\newblock
\APACrefYearMonthDay{2025}{}{}.
\newblock
\APACrefbtitle {Bielik v3 Small: Technical Report.} {Bielik v3 small: Technical
  report.}
\newblock
\begin{APACrefURL} {https://arxiv.org/abs/2505.02550} \end{APACrefURL}
\PrintBackRefs{\CurrentBib}

\bibitem [\protect \citeauthoryear {%
Ociepa%
, Flis%
, Wr{\'o}bel%
, Gwo{\'z}dziej%
\BCBL {}\ \BBA {} Kinas%
}{%
Ociepa%
\ \protect \BOthers {.}}{%
{\protect \APACyear {2024}}%
}]{%
ociepa2024bielik}
\APACinsertmetastar {%
ociepa2024bielik}%
\begin{APACrefauthors}%
Ociepa, K.%
, Flis, {\L}.%
, Wr{\'o}bel, K.%
, Gwo{\'z}dziej, A.%
\BCBL {} Kinas, R.%
\end{APACrefauthors}%
\unskip\
\newblock
\APACrefYearMonthDay{2024}{}{}.
\newblock
\APACrefbtitle {Bielik 7{B} v0.1: A {P}olish Language Model --- Development,
  Insights, and Evaluation.} {Bielik 7{B} v0.1: A {P}olish language model ---
  development, insights, and evaluation.}
\newblock
\begin{APACrefURL} {https://arxiv.org/abs/2410.18565} \end{APACrefURL}
\PrintBackRefs{\CurrentBib}

\bibitem [\protect \citeauthoryear {%
Piotrowski%
}{%
Piotrowski%
}{%
{\protect \APACyear {2012}}%
}]{%
piotrowski2012historical}
\APACinsertmetastar {%
piotrowski2012historical}%
\begin{APACrefauthors}%
Piotrowski, M.%
\end{APACrefauthors}%
\unskip\
\newblock
\APACrefYear{2012}.
\newblock
\APACrefbtitle {Natural Language Processing for Historical Texts} {Natural
  language processing for historical texts}.
\newblock
\APACaddressPublisher{}{Springer}.
\PrintBackRefs{\CurrentBib}

\bibitem [\protect \citeauthoryear {%
{Project Apertus}%
}{%
{Project Apertus}%
}{%
{\protect \APACyear {2025}}%
}]{%
apertus2025}
\APACinsertmetastar {%
apertus2025}%
\begin{APACrefauthors}%
{Project Apertus}%
\end{APACrefauthors}%
\unskip\
\newblock
\APACrefYearMonthDay{2025}{}{}.
\newblock
\APACrefbtitle {Apertus: Democratizing Open and Compliant {LLMs} for Global
  Language Environments.} {Apertus: Democratizing open and compliant {LLMs} for
  global language environments.}
\newblock
\begin{APACrefURL} {https://arxiv.org/abs/2509.14233} \end{APACrefURL}
\PrintBackRefs{\CurrentBib}

\bibitem [\protect \citeauthoryear {%
Przepi{\'o}rkowski%
, G{\'o}rski%
, {\L}azi{\'n}ski%
\BCBL {}\ \BBA {} P{\k{e}}zik%
}{%
Przepi{\'o}rkowski%
\ \protect \BOthers {.}}{%
{\protect \APACyear {2010}}%
}]{%
przepiorkowski2010nkjp}
\APACinsertmetastar {%
przepiorkowski2010nkjp}%
\begin{APACrefauthors}%
Przepi{\'o}rkowski, A.%
, G{\'o}rski, R.L.%
, {\L}azi{\'n}ski, M.%
\BCBL {} P{\k{e}}zik, P.%
\end{APACrefauthors}%
\unskip\
\newblock
\APACrefYearMonthDay{2010}{}{}.
\newblock
{\BBOQ}\APACrefatitle {Recent Developments in the {N}ational {C}orpus of
  {P}olish} {Recent developments in the {N}ational {C}orpus of
  {P}olish}.{\BBCQ}
\newblock
 \APACrefbtitle {{Proceedings of the Seventh International Conference on
  Language Resources and Evaluation (LREC'10)}.} {{Proceedings of the Seventh
  International Conference on Language Resources and Evaluation (LREC'10)}.}
\newblock
\APACaddressPublisher{Valletta, Malta}{European Language Resources Association
  (ELRA)}.
\newblock
\begin{APACrefURL} {https://aclanthology.org/L10-1097/} \end{APACrefURL}
\PrintBackRefs{\CurrentBib}

\bibitem [\protect \citeauthoryear {%
{Qwen Team}%
}{%
{Qwen Team}%
}{%
{\protect \APACyear {2026}}%
}]{%
qwen2026qwen38}
\APACinsertmetastar {%
qwen2026qwen38}%
\begin{APACrefauthors}%
{Qwen Team}%
\end{APACrefauthors}%
\unskip\
\newblock
\APACrefYearMonthDay{2026}{}{}.
\newblock
\APACrefbtitle {{Qwen3.8-2.4T-A95B}.} {{Qwen3.8-2.4T-A95B}.}
\newblock
\APAChowpublished {Model repository,
  \url{https://huggingface.co/Qwen/Qwen3.8-2.4T-A95B}}.
\PrintBackRefs{\CurrentBib}

\bibitem [\protect \citeauthoryear {%
Radford%
\ \protect \BOthers {.}}{%
Radford%
\ \protect \BOthers {.}}{%
{\protect \APACyear {2019}}%
}]{%
radford2019gpt2}
\APACinsertmetastar {%
radford2019gpt2}%
\begin{APACrefauthors}%
Radford, A.%
, Wu, J.%
, Child, R.%
, Luan, D.%
, Amodei, D.%
\BCBL {} Sutskever, I.%
\end{APACrefauthors}%
\unskip\
\newblock
\APACrefYearMonthDay{2019}{}{}.
\newblock
\APACrefbtitle {Language Models are Unsupervised Multitask Learners.} {Language
  models are unsupervised multitask learners.}
\newblock
\APAChowpublished {OpenAI technical report}.
\PrintBackRefs{\CurrentBib}

\bibitem [\protect \citeauthoryear {%
Raffel%
\ \protect \BOthers {.}}{%
Raffel%
\ \protect \BOthers {.}}{%
{\protect \APACyear {2020}}%
}]{%
raffel2020t5}
\APACinsertmetastar {%
raffel2020t5}%
\begin{APACrefauthors}%
Raffel, C.%
, Shazeer, N.%
, Roberts, A.%
, Lee, K.%
, Narang, S.%
, Matena, M.%
\BDBL {}Liu, P.J.%
\end{APACrefauthors}%
\unskip\
\newblock
\APACrefYearMonthDay{2020}{}{}.
\newblock
{\BBOQ}\APACrefatitle {Exploring the Limits of Transfer Learning with a Unified
  Text-to-Text Transformer} {Exploring the limits of transfer learning with a
  unified text-to-text transformer}.{\BBCQ}
\newblock
\APACjournalVolNumPages{Journal of Machine Learning Research}{21}{140}{1--67,}
\newblock

\newblock

\PrintBackRefs{\CurrentBib}

\bibitem [\protect \citeauthoryear {%
Sarkar%
}{%
Sarkar%
}{%
{\protect \APACyear {2024}}%
}]{%
sarkar2024storieslm}
\APACinsertmetastar {%
sarkar2024storieslm}%
\begin{APACrefauthors}%
Sarkar, S.%
\end{APACrefauthors}%
\unskip\
\newblock
\APACrefYearMonthDay{2024}{}{}.
\newblock
{\BBOQ}\APACrefatitle {{StoriesLM}: A Family of Language Models With
  Time-Indexed Training Data} {{StoriesLM}: A family of language models with
  time-indexed training data}.{\BBCQ}
\newblock
\APACjournalVolNumPages{SSRN Electronic Journal}{}{}{,}
\newblock
\begin{APACrefURL} {https://ssrn.com/abstract=4881024} \end{APACrefURL}
\newblock

\newblock

\PrintBackRefs{\CurrentBib}

\bibitem [\protect \citeauthoryear {%
Sennrich%
, Haddow%
\BCBL {}\ \BBA {} Birch%
}{%
Sennrich%
\ \protect \BOthers {.}}{%
{\protect \APACyear {2016}}%
}]{%
sennrich2016bpe}
\APACinsertmetastar {%
sennrich2016bpe}%
\begin{APACrefauthors}%
Sennrich, R.%
, Haddow, B.%
\BCBL {} Birch, A.%
\end{APACrefauthors}%
\unskip\
\newblock
\APACrefYearMonthDay{2016}{}{}.
\newblock
{\BBOQ}\APACrefatitle {Neural Machine Translation of Rare Words with Subword
  Units} {Neural machine translation of rare words with subword units}.{\BBCQ}
\newblock
 \APACrefbtitle {{Proceedings of the 54th Annual Meeting of the Association for
  Computational Linguistics (ACL)}.} {{Proceedings of the 54th Annual Meeting
  of the Association for Computational Linguistics (ACL)}.}
\PrintBackRefs{\CurrentBib}

\bibitem [\protect \citeauthoryear {%
Shazeer%
}{%
Shazeer%
}{%
{\protect \APACyear {2020}}%
}]{%
shazeer2020glu}
\APACinsertmetastar {%
shazeer2020glu}%
\begin{APACrefauthors}%
Shazeer, N.%
\end{APACrefauthors}%
\unskip\
\newblock
\APACrefYearMonthDay{2020}{}{}.
\newblock
\APACrefbtitle {{GLU} Variants Improve Transformer.} {{GLU} variants improve
  transformer.}
\newblock
\begin{APACrefURL} {https://arxiv.org/abs/2002.05202} \end{APACrefURL}
\PrintBackRefs{\CurrentBib}

\bibitem [\protect \citeauthoryear {%
Sheng%
, Chang%
, Natarajan%
\BCBL {}\ \BBA {} Peng%
}{%
Sheng%
\ \protect \BOthers {.}}{%
{\protect \APACyear {2019}}%
}]{%
sheng2019woman}
\APACinsertmetastar {%
sheng2019woman}%
\begin{APACrefauthors}%
Sheng, E.%
, Chang, K\BHBI W.%
, Natarajan, P.%
\BCBL {} Peng, N.%
\end{APACrefauthors}%
\unskip\
\newblock
\APACrefYearMonthDay{2019}{}{}.
\newblock
{\BBOQ}\APACrefatitle {The Woman Worked as a Babysitter: On Biases in Language
  Generation} {The woman worked as a babysitter: On biases in language
  generation}.{\BBCQ}
\newblock
 \APACrefbtitle {{Proceedings of the 2019 Conference on Empirical Methods in
  Natural Language Processing and the 9th International Joint Conference on
  Natural Language Processing (EMNLP-IJCNLP)}} {{Proceedings of the 2019
  Conference on Empirical Methods in Natural Language Processing and the 9th
  International Joint Conference on Natural Language Processing
  (EMNLP-IJCNLP)}}\ (\BPGS\ 3407--3412).
\PrintBackRefs{\CurrentBib}

\bibitem [\protect \citeauthoryear {%
Soldaini%
\ \protect \BOthers {.}}{%
Soldaini%
\ \protect \BOthers {.}}{%
{\protect \APACyear {2024}}%
}]{%
soldaini2024dolma}
\APACinsertmetastar {%
soldaini2024dolma}%
\begin{APACrefauthors}%
Soldaini, L.%
, Kinney, R.%
, Bhagia, A.%
, Schwenk, D.%
, Atkinson, D.%
, Authur, R.%
\BDBL {}Lo, K.%
\end{APACrefauthors}%
\unskip\
\newblock
\APACrefYearMonthDay{2024}{}{}.
\newblock
{\BBOQ}\APACrefatitle {{D}olma: An Open Corpus of Three Trillion Tokens for
  Language Model Pretraining Research} {{D}olma: An open corpus of three
  trillion tokens for language model pretraining research}.{\BBCQ}
\newblock
 \APACrefbtitle {{Proceedings of the 62nd Annual Meeting of the Association for
  Computational Linguistics (Volume 1: Long Papers)}} {{Proceedings of the 62nd
  Annual Meeting of the Association for Computational Linguistics (Volume 1:
  Long Papers)}}\ (\BPGS\ 15725--15788).
\PrintBackRefs{\CurrentBib}

\bibitem [\protect \citeauthoryear {%
Su%
\ \protect \BOthers {.}}{%
Su%
\ \protect \BOthers {.}}{%
{\protect \APACyear {2024}}%
}]{%
su2024roformer}
\APACinsertmetastar {%
su2024roformer}%
\begin{APACrefauthors}%
Su, J.%
, Ahmed, M.%
, Lu, Y.%
, Pan, S.%
, Bo, W.%
\BCBL {} Liu, Y.%
\end{APACrefauthors}%
\unskip\
\newblock
\APACrefYearMonthDay{2024}{}{}.
\newblock
{\BBOQ}\APACrefatitle {{RoFormer}: Enhanced Transformer with Rotary Position
  Embedding} {{RoFormer}: Enhanced transformer with rotary position
  embedding}.{\BBCQ}
\newblock
\APACjournalVolNumPages{Neurocomputing}{568}{}{,}
\newblock
\begin{APACrefDOI} \doi{10.1016/j.neucom.2023.127063} \end{APACrefDOI}
\newblock

\newblock

\PrintBackRefs{\CurrentBib}

\bibitem [\protect \citeauthoryear {%
Touvron%
\ \protect \BOthers {.}}{%
Touvron%
\ \protect \BOthers {.}}{%
{\protect \APACyear {2023}}%
}]{%
touvron2023llama}
\APACinsertmetastar {%
touvron2023llama}%
\begin{APACrefauthors}%
Touvron, H.%
, Lavril, T.%
, Izacard, G.%
, Martinet, X.%
, Lachaux, M\BHBI A.%
, Lacroix, T.%
\BDBL {}Lample, G.%
\end{APACrefauthors}%
\unskip\
\newblock
\APACrefYearMonthDay{2023}{}{}.
\newblock
\APACrefbtitle {{LLaMA}: Open and Efficient Foundation Language Models.}
  {{LLaMA}: Open and efficient foundation language models.}
\newblock
\begin{APACrefURL} {https://arxiv.org/abs/2302.13971} \end{APACrefURL}
\PrintBackRefs{\CurrentBib}

\bibitem [\protect \citeauthoryear {%
van Strien%
\ \protect \BOthers {.}}{%
van Strien%
\ \protect \BOthers {.}}{%
{\protect \APACyear {2020}}%
}]{%
vanstrien2020ocr}
\APACinsertmetastar {%
vanstrien2020ocr}%
\begin{APACrefauthors}%
van Strien, D.%
, Beelen, K.%
, Coll~Ardanuy, M.%
, Hosseini, K.%
, McGillivray, B.%
\BCBL {} Colavizza, G.%
\end{APACrefauthors}%
\unskip\
\newblock
\APACrefYearMonthDay{2020}{}{}.
\newblock
{\BBOQ}\APACrefatitle {Assessing the Impact of {OCR} Quality on Downstream
  {NLP} Tasks} {Assessing the impact of {OCR} quality on downstream {NLP}
  tasks}.{\BBCQ}
\newblock
 \APACrefbtitle {{Proceedings of the 12th International Conference on Agents
  and Artificial Intelligence (ICAART)}.} {{Proceedings of the 12th
  International Conference on Agents and Artificial Intelligence (ICAART)}.}
\PrintBackRefs{\CurrentBib}

\bibitem [\protect \citeauthoryear {%
Vaswani%
\ \protect \BOthers {.}}{%
Vaswani%
\ \protect \BOthers {.}}{%
{\protect \APACyear {2017}}%
}]{%
vaswani2017attention}
\APACinsertmetastar {%
vaswani2017attention}%
\begin{APACrefauthors}%
Vaswani, A.%
, Shazeer, N.%
, Parmar, N.%
, Uszkoreit, J.%
, Jones, L.%
, Gomez, A.N.%
\BDBL {}Polosukhin, I.%
\end{APACrefauthors}%
\unskip\
\newblock
\APACrefYearMonthDay{2017}{}{}.
\newblock
{\BBOQ}\APACrefatitle {Attention Is All You Need} {Attention is all you
  need}.{\BBCQ}
\newblock
 \APACrefbtitle {{Advances in Neural Information Processing Systems
  (NeurIPS)}.} {{Advances in Neural Information Processing Systems (NeurIPS)}.}
\PrintBackRefs{\CurrentBib}

\bibitem [\protect \citeauthoryear {%
Wojczulis%
\ \BBA {} K{\l}eczek%
}{%
Wojczulis%
\ \BBA {} K{\l}eczek%
}{%
{\protect \APACyear {2021}}%
}]{%
papugapt2}
\APACinsertmetastar {%
papugapt2}%
\begin{APACrefauthors}%
Wojczulis, M.%
\BCBT {}\ \BBA {} K{\l}eczek, D.%
\end{APACrefauthors}%
\unskip\
\newblock
\APACrefYearMonthDay{2021}{}{}.
\newblock
\APACrefbtitle {{papuGaPT2}: Polish {GPT2} Language Model.} {{papuGaPT2}:
  Polish {GPT2} language model.}
\newblock
\APAChowpublished {Model repository,
  \url{https://huggingface.co/flax-community/papuGaPT2}}.
\PrintBackRefs{\CurrentBib}

\bibitem [\protect \citeauthoryear {%
Yazar%
, Kutlu%
\BCBL {}\ \BBA {} Bay{\i}rl{\i}%
}{%
Yazar%
\ \protect \BOthers {.}}{%
{\protect \APACyear {2025}}%
}]{%
yazar2025turkronicles}
\APACinsertmetastar {%
yazar2025turkronicles}%
\begin{APACrefauthors}%
Yazar, T.%
, Kutlu, M.%
\BCBL {} Bay{\i}rl{\i}, {\.I}.K.%
\end{APACrefauthors}%
\unskip\
\newblock
\APACrefYearMonthDay{2025}{}{}.
\newblock
{\BBOQ}\APACrefatitle {Turkronicles: Diachronic Resources for the Fast Evolving
  {T}urkish Language} {Turkronicles: Diachronic resources for the fast evolving
  {T}urkish language}.{\BBCQ}
\newblock
\APACjournalVolNumPages{Language Resources and Evaluation}{59}{4}{3765--3797,}
\newblock
\begin{APACrefDOI} \doi{10.1007/s10579-025-09857-w} \end{APACrefDOI}
\newblock

\newblock

\PrintBackRefs{\CurrentBib}

\bibitem [\protect \citeauthoryear {%
{Z.ai}%
}{%
{Z.ai}%
}{%
{\protect \APACyear {2026}}%
}]{%
glm2026glm52}
\APACinsertmetastar {%
glm2026glm52}%
\begin{APACrefauthors}%
{Z.ai}%
\end{APACrefauthors}%
\unskip\
\newblock
\APACrefYearMonthDay{2026}{}{}.
\newblock
\APACrefbtitle {{GLM-5.2}.} {{GLM-5.2}.}
\newblock
\APAChowpublished {Model repository,
  \url{https://huggingface.co/zai-org/GLM-5.2}}.
\PrintBackRefs{\CurrentBib}

\bibitem [\protect \citeauthoryear {%
Zhang%
\ \BBA {} Sennrich%
}{%
Zhang%
\ \BBA {} Sennrich%
}{%
{\protect \APACyear {2019}}%
}]{%
zhang2019rmsnorm}
\APACinsertmetastar {%
zhang2019rmsnorm}%
\begin{APACrefauthors}%
Zhang, B.%
\BCBT {}\ \BBA {} Sennrich, R.%
\end{APACrefauthors}%
\unskip\
\newblock
\APACrefYearMonthDay{2019}{}{}.
\newblock
{\BBOQ}\APACrefatitle {Root Mean Square Layer Normalization} {Root mean square
  layer normalization}.{\BBCQ}
\newblock
 \APACrefbtitle {{Advances in Neural Information Processing Systems
  (NeurIPS)}.} {{Advances in Neural Information Processing Systems (NeurIPS)}.}
\PrintBackRefs{\CurrentBib}

\bibitem [\protect \citeauthoryear {%
Zheng%
\ \protect \BOthers {.}}{%
Zheng%
\ \protect \BOthers {.}}{%
{\protect \APACyear {2023}}%
}]{%
zheng2023judging}
\APACinsertmetastar {%
zheng2023judging}%
\begin{APACrefauthors}%
Zheng, L.%
, Chiang, W\BHBI L.%
, Sheng, Y.%
, Zhuang, S.%
, Wu, Z.%
, Zhuang, Y.%
\BDBL {}Stoica, I.%
\end{APACrefauthors}%
\unskip\
\newblock
\APACrefYearMonthDay{2023}{}{}.
\newblock
{\BBOQ}\APACrefatitle {Judging {LLM}-as-a-Judge with {MT}-Bench and Chatbot
  Arena} {Judging {LLM}-as-a-judge with {MT}-bench and chatbot arena}.{\BBCQ}
\newblock
 \APACrefbtitle {{Advances in Neural Information Processing Systems (NeurIPS),
  Datasets and Benchmarks Track}.} {{Advances in Neural Information Processing
  Systems (NeurIPS), Datasets and Benchmarks Track}.}
\PrintBackRefs{\CurrentBib}

\end{thebibliography}

\end{document}